\pdfoutput=1

\documentclass[11pt]{article}

\usepackage[final]{acl}

\usepackage{times}
\usepackage{latexsym}

\usepackage{multirow}
\usepackage[T1]{fontenc}
\usepackage{times}
\usepackage{latexsym}

\usepackage{multirow}
\usepackage{booktabs}
\usepackage{tablefootnote}

\usepackage[utf8]{inputenc}

\usepackage{microtype}

\usepackage{inconsolata}

\usepackage{graphicx}

\title{INDIC QA BENCHMARK: A Multilingual Benchmark to Evaluate Question Answering capability of LLMs for Indic Languages}

\author{
  Abhishek kumar singh$^\diamond$, Vishwajeet Kumar$^\S$, Rudra Murthy$^\S$\
  \\
 \textbf{Jaydeep Sen$^\S$, Ashish Mittal$^\S$, Ganesh Ramakrishnan$^\diamond$}
  \\
  $^\diamond$Indian Institute of Technology Bombay, India
  \\
  $^\S$IBM Research, India
  \\
  \texttt{\{abhisheksingh,ganesh\}@cse.iitb.ac.in}, 
  \\
  \texttt{\{vishk024,rmurthyv,jaydesen,arakeshk\}@in.ibm.com}
}

\begin{document}
\maketitle
\begin{abstract}
Large Language Models (LLMs) perform well on unseen tasks in English, but their abilities in non-English languages are less explored due to limited benchmarks and training data. To bridge this gap, we introduce the Indic-QA Benchmark, a large dataset for context-grounded question answering in 11 major Indian languages, covering both extractive and abstractive tasks. Evaluations of multilingual LLMs, including instruction fine-tuned versions, revealed weak performance in low-resource languages due to a strong English-language bias in their training data. We also investigated the Translate-Test paradigm,where inputs are translated to English for processing and the results are translated back into the source language for output. This approach outperformed multilingual LLMs, particularly in low-resource settings. By releasing Indic-QA, we aim to promote further research into LLMs’ question-answering capabilities in low-resource languages. This benchmark offers a critical resource to address existing limitations and foster multilingual understanding.
\footnote{Source code and Data are available at \url{https://github.com/ayushayush591/IndicQA-Benchmark}.}
\end{abstract}

\begin{figure}[h!]
\centering
\begin{minipage}{0.48\textwidth}
  \centering
  \includegraphics[trim={1.5cm 0cm 2cm 0.5cm}, clip, width=\linewidth]{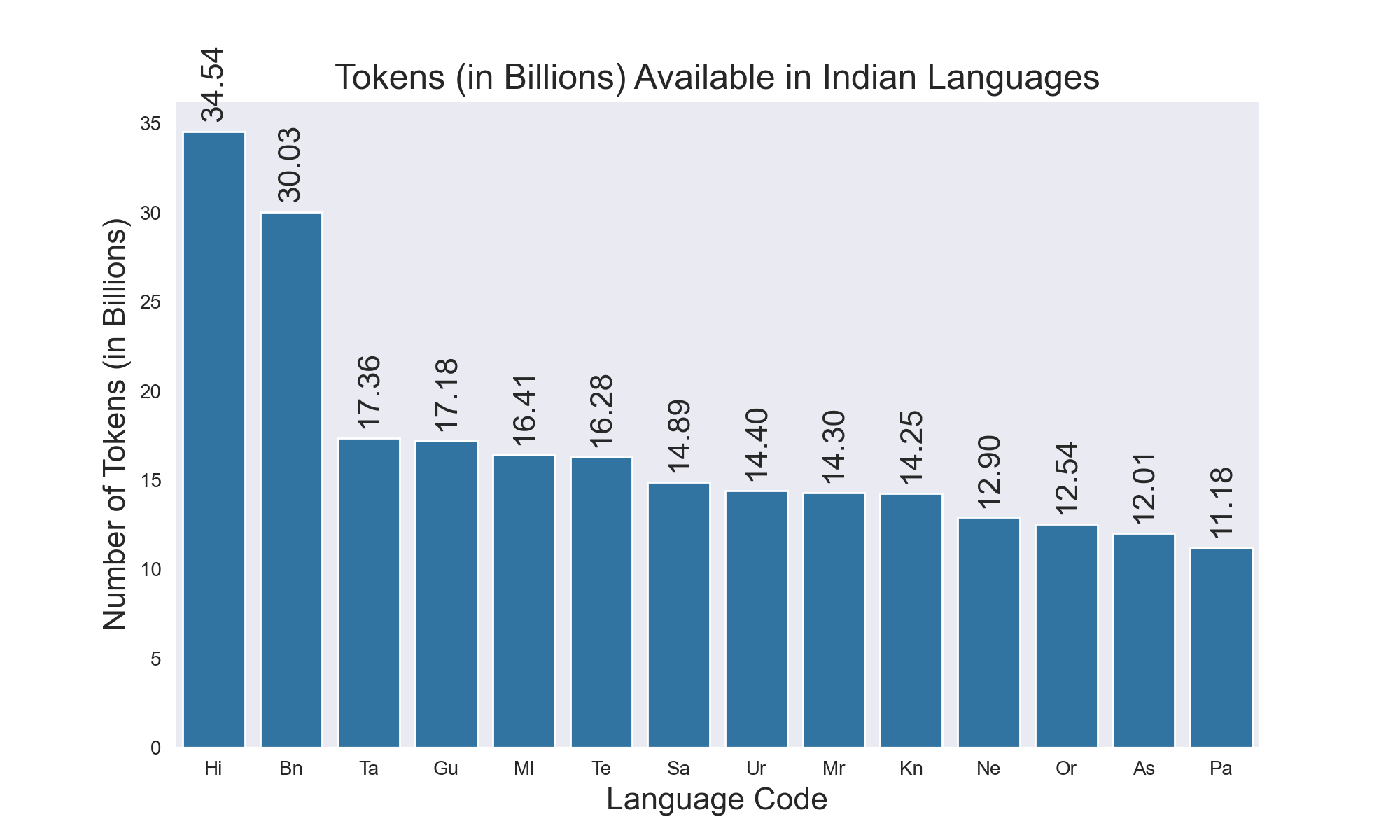}
  \caption{Total tokens available for each Indian language in the Sangraha Data \cite{rahman2024indicllmsuite}. In contrast, RefinedWeb \cite{penedo2023refinedwebdatasetfalconllm} contains around 5 Trillion tokens in English.}
  \label{fig:sangraha_stas}
\end{minipage}
\hfill
\begin{minipage}{0.48\textwidth}
  \centering
  \includegraphics[width=\linewidth]{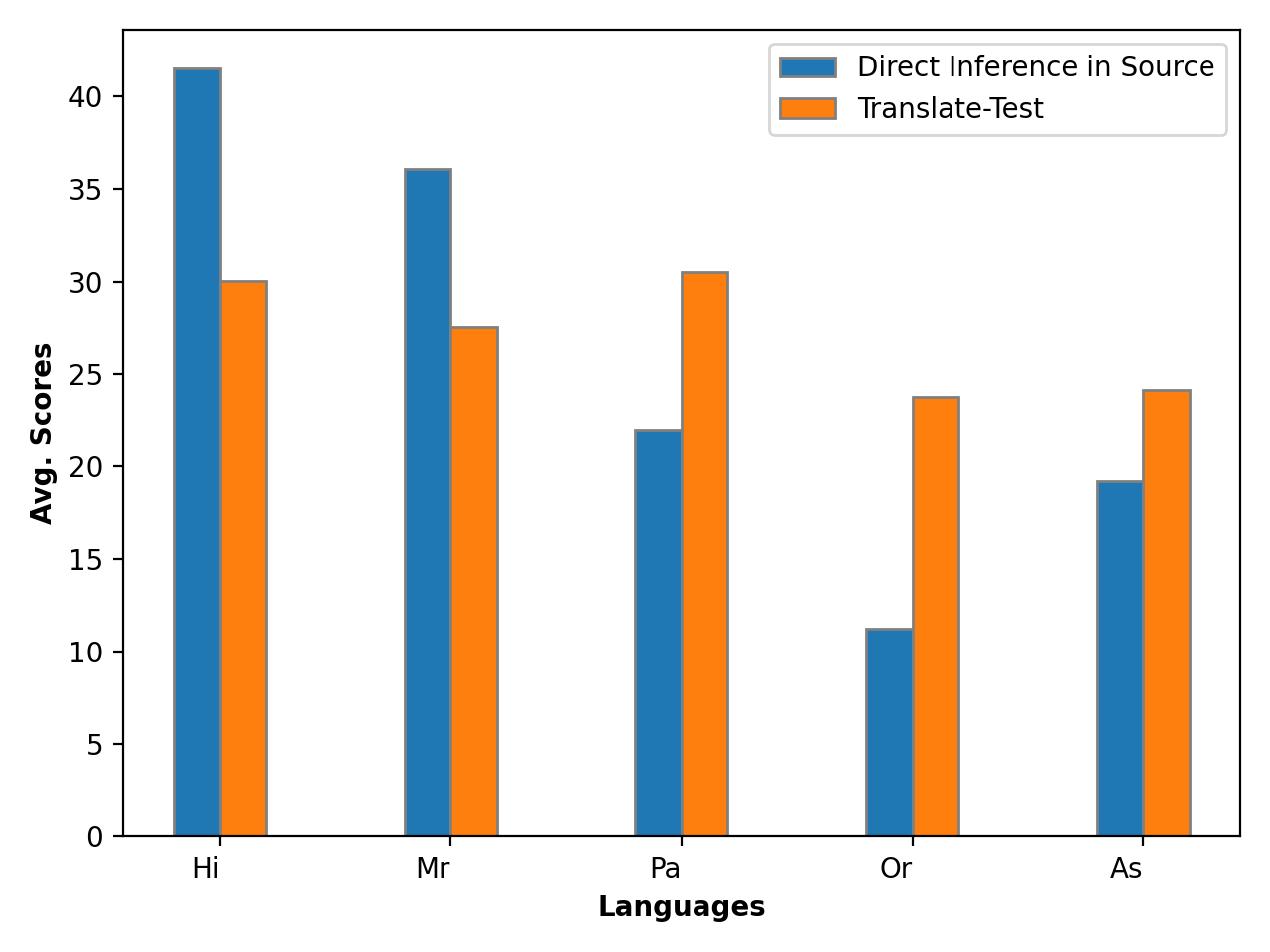}
  \caption{Comparison of \textbf{LLama 3-8B} evaluation results using source language test set vs Translate test set. The results clearly indicate that the Translate-Test paradigm yields better scores for low-resource languages (Punjabi, Odia, Assamese), whereas the source language test set gets better scores for mid-resource language Hindi and Marathi which has high correlation with Hindi.}
  \label{fig:Comparison_graph}
\end{minipage}
\end{figure}

India, with a population of almost 1.4 billion people, is home to numerous major languages that are considered low-resource by the natural language processing (NLP) community. Despite the growing capabilities of Large Language Models (LLMs) in tasks like context-grounded question answering (CQA) in English, their performance in non-English languages remains underexplored due to a lack of high-quality datasets. To address this gap, we introduce Indic-QA, the largest publicly available context-grounded question-answering dataset for 11 major Indian languages from two language families. This dataset encompasses both extractive and abstractive QA tasks, incorporating existing datasets as well as English QA datasets translated into Indian languages. Additionally, we generate a synthetic dataset using the Gemini model, with manual verification for quality assurance.Our evaluation of various multilingual LLMs and their instruction-fine-tuned variants on the Indic-QA benchmark reveals subpar performance, particularly for low-resource languages. This outcome highlights the English language bias inherent in these models due to predominantly English pre-training data. 

We tested the Translate-Test paradigm as an alternative, which translates input from the source language to English, utilizes the LLM's problem-solving ability in English, and then translates the response back to the source language. Our investigation shows that while multilingual LLMs perform better in mid-resource languages, the Translate-Test paradigm significantly outperforms them in low-resource languages. \\ \textbf{What distinguishes our benchmark from other existing Multilingual Indic context-grounded question-answering Benchmarks?} 
There are numerous context-grounded question-answering benchmarks available for high-resource languages like English. However, there are very few benchmarks available for Indic languages[\ref{rel_work}], and those that do exist often lack domain diversity and are limited in size. To address these gaps, we developed the Indic-QA benchmark. We sampled a variety of Wikipedia and Common Crawl pages, focusing on paragraphs rich in cultural nuances, to create a comprehensive and culturally diverse benchmark for Indic languages. Our findings show that base pre-trained models frequently produce incorrect or illogical answers. However, few-shot prompting notably improves answer quality by guiding the models to extract more precise information from the text. This paper makes the following key contributions:
\begin{enumerate}
\item \textbf{Indic-QA BENCHMARK}: We release a multilingual evaluation benchmark for assessing the Indic Question-Answering capabilities of LLMs, focusing on low-resource languages and multi-domain abstractive tasks.
\item \textbf{Empirical Evaluation}: We critically evaluate several esteemed LLMs for Indic languages, comparing their performance on the new benchmark to determine their QA skills.
\item \textbf{Translate-Test Paradigm}: We conduct an empirical study using the Translate-Test approach and direct generation in source languages with multilingual LLMs, demonstrating that the Translate-Test approach offers competitive and often superior performance for low-resource languages.
\end{enumerate}

\begin{table*}[!h]
\centering
\label{ref:dataset}
\resizebox{\textwidth}{!}{%
\begin{tabular}{lrrrrrrrrrrrr} 
\toprule
\multicolumn{1}{c}{\textbf{Datasets}} & \multicolumn{1}{c}{\textbf{As}} & \multicolumn{1}{c}{\textbf{Bn}} & \multicolumn{1}{c}{\textbf{Gu}} & \multicolumn{1}{c}{\textbf{Hi}} & \multicolumn{1}{c}{\textbf{Kn}} & \multicolumn{1}{c}{\textbf{Ml}} & \multicolumn{1}{c}{\textbf{Mr}} & \multicolumn{1}{c}{\textbf{Od}} & \multicolumn{1}{c}{\textbf{Pa}} & \multicolumn{1}{c}{\textbf{Ta}} & \multicolumn{1}{c}{\textbf{Te}} \\ 
\hline     
Hindi Squad & 3099 & 3107 & 3371 & \underline{4734} & 3068 & 2926 & 3165 & 3079 & 3469 & 2743 & 2955 \\
NQ Open & 1462 & 1483 & 1570 & 1842 & 1447 & 1420 & 1511 & 1451 & 1570 & 1331 & 1420 \\
Chaii & 339 & 351 & 394 & \underline{746} & 351 & 328 & 373 & 305 & 388 & 325 & 361 \\
Indic QA & \underline{1789} & \underline{1763} & 1369 & \underline{1547} & \underline{1517} & \underline{1589} & \underline{1604} & \underline{1680} & \underline{1542} & \underline{1804} & \underline{1734} \\
XSquad & \underline{1190} & \underline{1190} & \underline{1190} & \underline{1190} & \underline{1190} & \underline{1190} & \underline{1190} & \underline{1190} & \underline{1190} & \underline{1190} & \underline{1190}\\
XORQA & 537 & 538 & 532 & 537 & 534 & 533 & 529 & 529 & 531 & 537 & 538 \\
MLQA & 2362 & 2403 & 2718 & \underline{4918} & 2299 & 2128 & 2433 & 2370 & 2730 & 2129 & 2291 \\
\midrule
Synthetic MCQA$^{*}$ & 1741 & 1662 & 2162 & 3802 & 1618 & 1248 & 1807 & 1753 & 2326 & 1150 & 1416 \\
MS Marco$^{*}$ & 29724 & 30089 & 31741 & 35735 & 29212 & 28528 & 30180 & 30073 & 32032 & 27197 & 28995 \\
Llama Index$^{*}$ & 1158 & 1312 & 1333 & 1384 & 1310 & 1250 & 1316 & 1263 & 1175 & 1258 & 1306 \\
\bottomrule
\end{tabular}%
}
\caption{Indic-QA Dataset Statistics. Indic-QA benchmark is a compilation of existing datasets, English datasets translated to Indian languages, and, synthetic dataset generated using Gemini. The dataset comprises Extractive Question Answering and abstractive Question Answering (${*}$). As: Assamese, Bn: Bengali, Gu: Gujarati, Hi: Hindi, Kn: Kannada, Ml:Malayalam, Mr: Marathi, Od: Odia, Pa: Punjabi, Ta: Tamil, Te: Telugu. We have translated NQ Open, XORQA, Llama Index, and MS Marco datasets to Hindi. We have translated all the above datasets to the remaining ten Indian languages (underline data instances were already present in the referred language.)}
\label{ref:data_table}
\end{table*}

\section{Related work} \label{rel_work}
In the realm of context-grounded question answering (QA), significant research has been conducted in both English and Indian languages. This task involves presenting a question along with a contextual paragraph to the model, which then extracts the phrase from the paragraph. Various benchmarks \cite{dzendzik2021english,rajpurkar2016squad,rajpurkar2018know} have been established for this task, with encoder-only transformer models proving effective in extracting the span containing the answer from the paragraph. The Indic QA community has demonstrated remarkable performance using models like XLM-RoBERTa\cite{conneau2019unsupervised} and others, particularly for multilingual Indian languages. They have a rich dataset to showcase their benchmarks, including SQuAD \cite{rajpurkar2016squad} for English, along with its translated version in Hindi. Additionally, instead of translation, there are datasets specifically designed for evaluating benchmarks in Hindi, such as the Chaii \footnote{\url{https://www.kaggle.com/competitions/chaii-hindi-and-tamil-question-answering}}.
 dataset and IndicQA, which are also discussed in this survey paper \cite{kolhatkar2023indic}. Although there are a few benchmarks for Indic Question Answering, they lack extensive domain coverage, which is crucial for evaluating the robustness of models. In contrast, English benchmarks encompass a wide range of domain-specific datasets such as Resources like the llama Index \cite{Liu_LlamaIndex_2022} highlight that selecting the appropriate evaluation dataset is challenging and highly dependent on the specific use case. Academic benchmarks such as BEIR\cite{thakur2021beir} and HotpotQA\cite{yang2018hotpotqa} often fail to generalize across different use cases. For example, parameters that work well on certain data domains (e.g., SEC filings) may not perform as effectively on others (e.g., research papers). This challenge led them to the create a dataset hub specifically designed for evaluating RAG systems, encompassing a wide range of domains including research papers, blockchain articles, and code articles.

Additionally, NQ Open \cite{lee-etal-2019-latent} contains a wealth of Wikipedia content across various domains, and MS MARCO \cite{DBLP:journals/corr/abs-2108-13897} features questions sampled from real-world user searches with contexts derived from web documents. The diversity of user queries leads to a broad range of content, making MS MARCO highly versatile. Although initially intended for a different task, we adopted this dataset for our purposes.
Hence to address the lack of existing Indic QA benchmark datasets, we translated and adapted several commonly used English QA datasets into 11 Indic languages. This approach provides a more comprehensive and robust evaluation framework for Indic Question Answering models. By leveraging these datasets, we aim to offer a diverse and extensive evaluation resource, enhancing the development and assessment of QA models in Indic languages.

Earlier attempts at the Translate-Test approach \cite{etxaniz2023multilingual,intrator2024breaking} faced limitations due to less advanced translation systems. However, the emergence of larger bilingual parallel datasets \cite{reid2022role} has allowed researchers to develop robust neural translation models, greatly improving translation performance. To the best of our knowledge, no one has Tried the Translate-Test approach for Indic QA systems.

\section{Benchmarks}
The primary focus of this work is on context-based QA, where the answer is found within the given context. The datasets utilized in this study were tailored to facilitate this task, with each instance composed of triples consisting of a context, a question, and an answer. This section provides a detailed description of the methodology used to create or modify the existing dataset for our task.

\begin{figure*}[h!]
    \centering
    \includegraphics[width=0.75\linewidth]{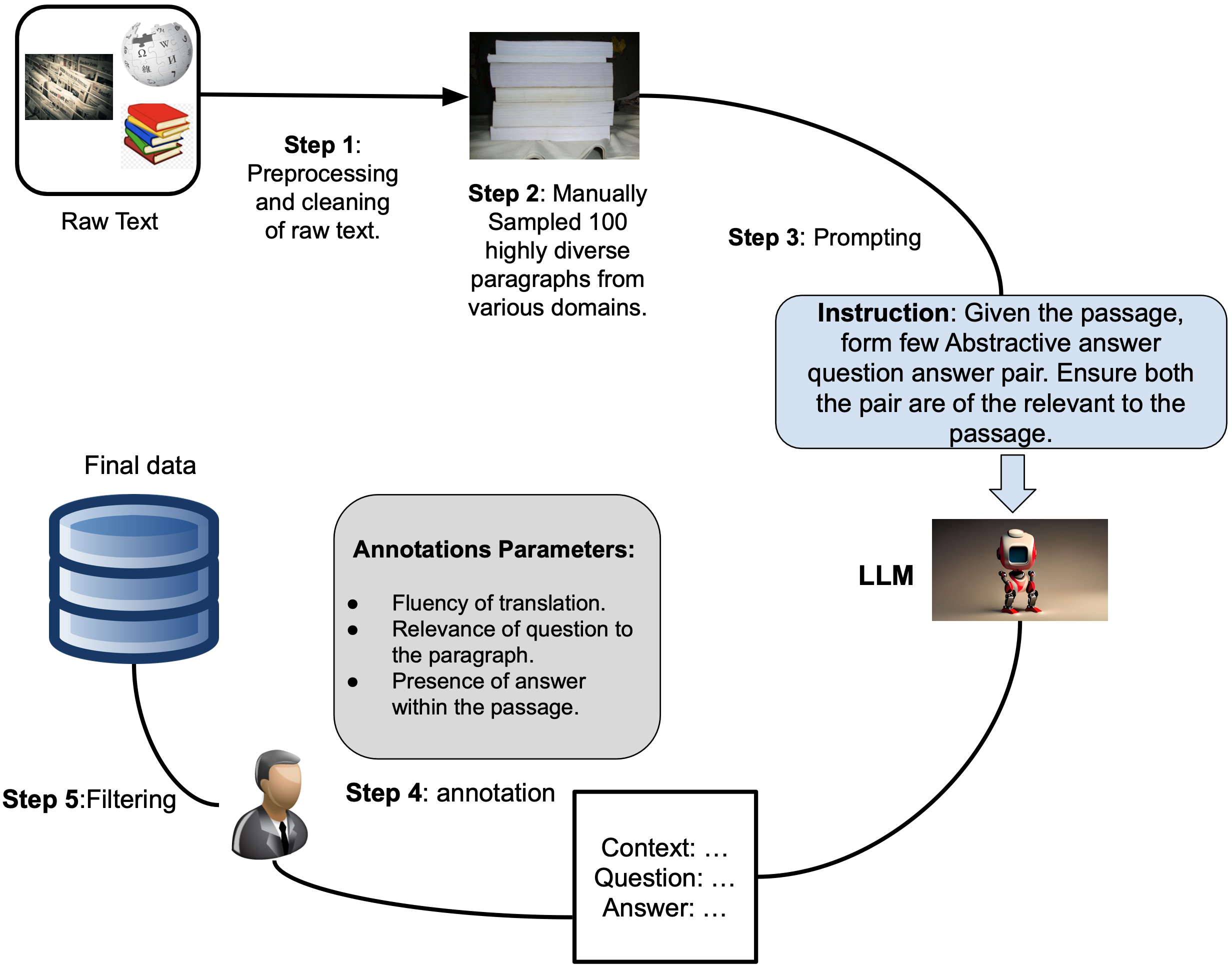}
    \caption{Workflow of synthetic data creation, in the fig. LLM used is Gemini-pro model.}
    \label{fig:workflow-synthetic}
    \end{figure*}
    
\subsection{Datasets} \label{dataset_section}
In this section, we provide a catalog of the datasets constituting this benchmark, complete with a thorough exposition of their original accessibility and the modifications we have implemented. These datasets are either pre-existing or have been released as part of this work. Following is a detailed description of each dataset.


\begin{enumerate}
    \item \textbf{Hindi \texttt{SQuAD}}: This dataset is a translated version of the original \texttt{SQuAD} \cite{rajpurkar2016squad} into Hindi. It consists of nearly 5,000 instances, translated using the Google Translate API. We translated that from Hindi to other Indic languages.
    \item \textbf{X\texttt{QuAD}}: \texttt{XQuAD} (Cross-lingual Question Answering Dataset) \cite{artetxe2019cross} is a benchmark for evaluating cross-lingual question answering performance. It consists of 240 paragraphs and 1,190 question-answer pairs sourced from the SQuAD v1.1 development set \cite{rajpurkar2016squad}, with professional translations into ten languages. However, we use the version from \cite{singh2024indicgenbench}, which includes manual translations for all Indic languages.
    \item \textbf{ChaII Dataset} \cite{thirumala2022extractive}: This question-answering dataset features context-question-answer triples in Hindi and Tamil, gathered directly without translation. Created by expert data annotators who are native speakers, the dataset presents a realistic information-seeking task focused on predicting answers to genuine questions about Wikipedia articles. It was used in a Kaggle challenge and includes 1104 questions in Hindi and Tamil, we used the Hindi part of the data and translated it to 10 other Indian languages.
    
    \item \textbf{Indic QA} \cite{Doddapaneni2022towards}: This dataset is a manually curated cloze-style reading comprehension dataset designed for evaluating question-answering models in 10 Indic languages Since this dataset doesn't have Gujarati translation we translated it from Hindi to Gujarati and validated the translation as described in the [\ref{methodology}] section.
    
    \item \textbf{MLQA} \cite{lewis2019mlqa}: \texttt{MLQA} (MultiLingual Question Answering) is a benchmark dataset for evaluating cross-lingual question answering performance. We have used the \texttt{MLQA} test set for benchmarking purposes, the test set contains 4918 triples of the form (context, question, answer) all available in Hindi, hence we translated this triplet from Hindi to 10 other Indian languages.

    \item \textbf{MS Marco} \cite{DBLP:journals/corr/abs-2108-13897}: Microsoft Machine Reading Comprehension (MS MARCO) is a collection of large-scale datasets designed for deep learning applications related to search. The questions in MS Marco are sampled from real, anonymized user queries. The context passages, from which the answers are derived, are extracted from real web documents using the latest version of the Bing search engine. We initially considered adapting the multilingual version of the MS MARCO passage ranking dataset (mMarco) for our setting. However, since mMarco lacks a test set, we opted to use the MS MARCO test set, which contains 100k instances, each including a query and a set of passages, among which only one is relevant to the query. We filtered out instances without any relevant passages, resulting in a dataset of 55k instances.
    
    We then translated this dataset from English to Hindi. After applying certain filtering conditions, The exact steps are detailed in [\ref{methodology}]. The final dataset now includes the question, the source document, and the corresponding answer, and is available in 11 Indian languages.
    
    \item \textbf{NQ-Open trans} \cite{lee-etal-2019-latent}: The \texttt{NQ-Open} task is an open-domain question-answering benchmark derived from Natural Questions. The objective is to predict an English answer string for a given English question, with all questions answerable using the contents of English Wikipedia. Initially, the dataset was entirely in English, with context, question, and answer all in English. The context often included tables scraped from HTML pages of Wikipedia, resulting in numerous HTML tags. To clean the dataset, we removed all triples where the context contained a table and eliminated all other HTML tags from the remaining examples. In this modified dataset, the fields include the source document (the entire Wikipedia page), the long answer (a paragraph from the page containing the answer), and the exact phrase or word from that paragraph as the short answer. We modified the long answer to serve as the context and the short answer as the answer for the corresponding question. and Since after all this modification dataset was in English we translated that to other Indian languages. 

    \item \textbf{XORQA} \cite{asai2020xor}: 
    Cross-lingual Open Retrieval Question Answering (XOR QA) consists of three tasks involving cross-lingual document retrieval from both multilingual and English resources. This dataset was subsequently translated into other Indian languages by \cite{singh2024indicgenbench}. We utilized the same since it was cross-lingual data, the context was in English while the questions and answers were in other languages. To adapt it to our setting, we translated the context into various Indian languages.
    
    \item \textbf{LLama Index} \footnote{\url{https://www.llamaindex.ai/blog/introducing-llama-datasets-aadb9994ad9e}} \cite{Liu_LlamaIndex_2022}: 
    The dataset includes question-answer pairs along with source context, serving as an evaluation tool for the RAG pipeline. We observed that some contexts were insufficient to answer the questions effectively. To address this, we applied the BGE-M3 \cite{bge-m3} algorithm to measure the similarity between the context and the query, using a threshold of 0.43 to determine if a question could be answered adequately based on the context. Post filtering we translated the resulting context, question, and answer triplets into Hindi and Hindi other Indian languages.
    
    
    \item \textbf{Synthetic Data}:
    This dataset is introduced as part of this study. We employed the Gemini model \cite{team2023gemini} to generate question-answer pairs based on provided contexts. To achieve this, we sampled a diverse set of Hindi contexts from sources such as Wikipedia, storybooks, Indian news articles, and paragraphs from competitive exams. We then prompted the model with these context paragraphs to generate abstractive question-answer pairs, framing the task as a abstractive QA task.
    Subsequently, this dataset was translated into other languages and verified by language experts, the whole workflow process can be found [\ref{fig:workflow-synthetic}] [\ref{syn_data}].

\end{enumerate}

\begin{figure*}[h!]
    \centering
    \includegraphics[width=1.0\linewidth]{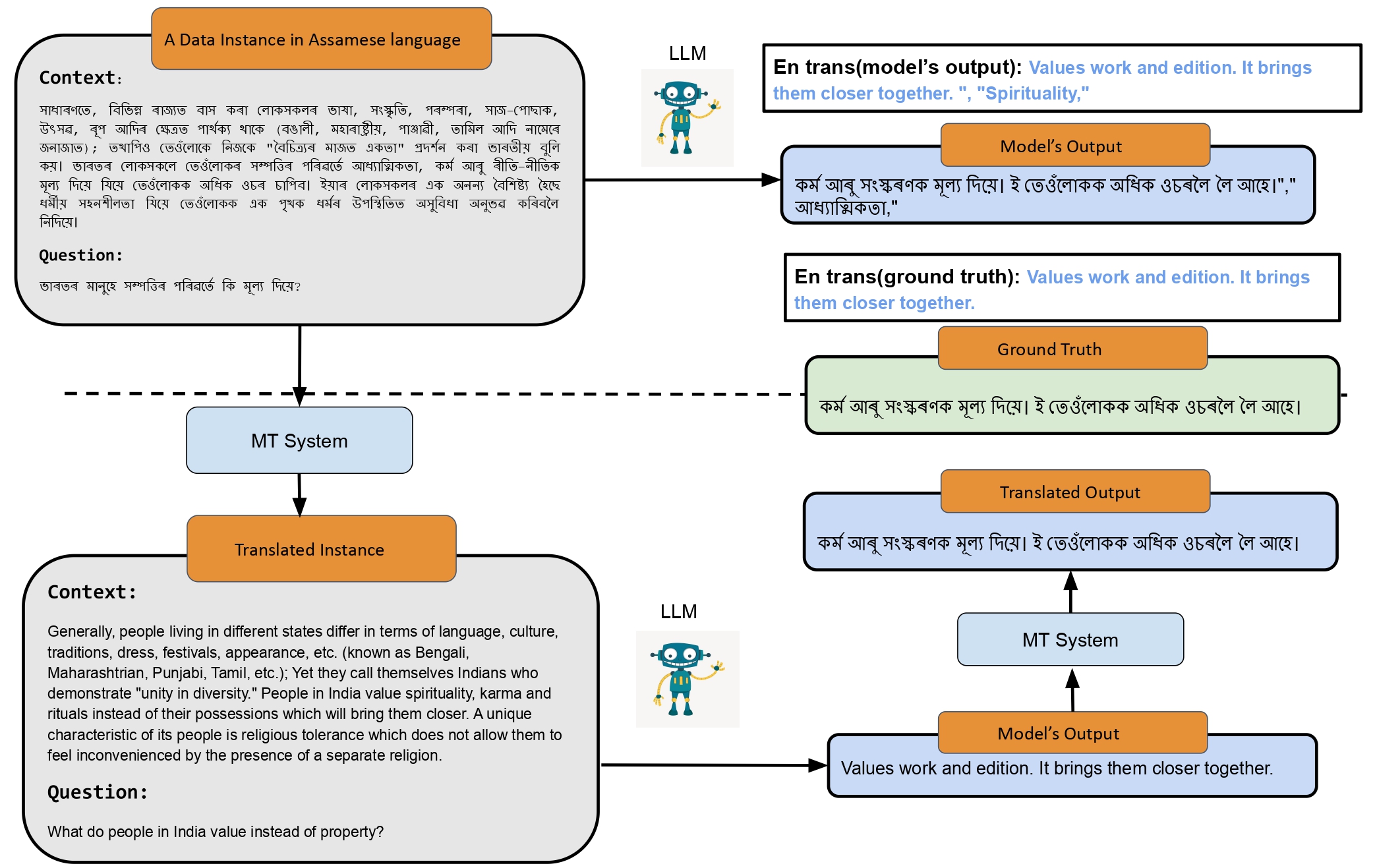}
    \caption{Inference in source Language (Top) vs Translate Test Inference (Bottom).}
    \label{fig:res_summary}
    \end{figure*}
    
\subsection{Data Curation Methodology} \label{methodology}
In light of the approaches discussed previously in Section \ref{rel_work}, context-grounded question-answering datasets can generally be categorized into two types: abstractive and extractive. Although many extractive datasets exist for high-resource languages, the few available for Indian languages lack diversity in domains and question types, limiting their usefulness for benchmarking. Hence, we extended the benchmark suite available in English to these Indian languages by translating. We utilized IndicTrans2\footnote{\url{https://github.com/AI4Bharat/IndicTrans2}} \cite{gala2023indictrans2} for translation, an open-source transformer-based multilingual NMT model that supports high-quality translations across all the 22 scheduled Indian languages. We segmented the context paragraph into sentences using the Spacy library, translated each sentence, and then recombined them. This approach yielded better translation results, and importantly, the model did not lose context when translating, thus preserving the coherence of the text. In the list of datasets for benchmarking, some are available only in English (e.g., NQ-open, ORQA, llama index,MS-Marco), while others are available in both English and Hindi (e.g., Hindi SQuAD, CHAII, MLQA, Synthetic data). Additionally, a few datasets (e.g., IndicQA, XSQuAD) are also available in all 10 or 11 languages with verified translations. For all the datasets not found in the respective language, we translated them and applied the filtering methods discussed below.

To assess the quality of our translations, we initially translated each dataset from the source language to the target language, followed by back-translation from the target language back to the source language. We then calculated the CHRF scores\cite{popovic2015chrf} between the original and back-translated sentences, using a threshold of 50 to filter the instances. Additionally, we manually verified a subset of the filtered data to ensure accuracy. For the translation process, we initially translated the English data directly into Hindi. After filtering the data, we then translated it from Hindi to other Indian languages, rather than directly from English. This approach was based on our observation that the translation quality from Hindi to other Indian languages was superior. The improved quality can be attributed to the linguistic similarities within the same language family, including morphology, syntax, and grammar.

\begin{table*}[!h]
\centering
\resizebox{1.0\linewidth}{!}{%
\begin{tabular}{crrrrrrrrrrr} 
\toprule
\textbf{Languages} & \multicolumn{2}{c}{Bloom} & \multicolumn{2}{c}{Gemma} & \multicolumn{2}{c}{Llama-3} & \multicolumn{2}{c}{Llama3.1} & \multicolumn{2}{c}{Gemma-2} \\
\cmidrule(lr){1-1} \cmidrule(lr){2-3} \cmidrule(lr){4-5} \cmidrule(lr){6-7} \cmidrule(lr){8-9} \cmidrule(lr){10-11}
\multicolumn{1}{l}{} & \begin{tabular}[c]{@{}c@{}}Direct\\ Inference\end{tabular} & \multicolumn{1}{c}{Translate} & \begin{tabular}[c]{@{}c@{}}Direct\\ Inference\end{tabular} & \multicolumn{1}{c}{Translate} & \begin{tabular}[c]{@{}c@{}}Direct\\ Inference\end{tabular} & \multicolumn{1}{c}{Translate} & \begin{tabular}[c]{@{}c@{}}Direct\\ Inference\end{tabular} & \multicolumn{1}{c}{Translate} & \begin{tabular}[c]{@{}c@{}}Direct\\ Inference\end{tabular} & \multicolumn{1}{c}{Translate} \\
\midrule
\textbf{As} & 13.86  & \textbf{16.19} & 21.70 & 16.08 & 19.23 & \textbf{24.14} & 21.29 & \textbf{22.48} & 33.01 & \textbf{35.60} \\
\textbf{Bn} & 17.84 & 16.07 & 21.15 & 16.69 & 19.14 & \textbf{23.96} & 27.61 & 24.00 & 38.33 & 35.11\\
\textbf{Gu} & 13.27 & \textbf{18.59} & 24.90 & \textbf{25.12} & 20.27 & \textbf{29.55} & 24.08 & \textbf{32.11} & 40.10 & \textbf{44.86}\\
\textbf{Hi} & 21.69 & 19.29 & 34.37 & 19.72 & 41.53 & 30.02 & 46.76 & 26.22 & 44.46 & 43.18\\
\textbf{Kn} & 15.63 & \textbf{17.86} & 24.47 & 16.31 & 20.39 & \textbf{25.84} & 21.31 & \textbf{26.22} & 35.31 & \textbf{37.98}\\
\textbf{Ml} & 19.22 & 17.96 & 25.20 & 17.31 & 22.80 & \textbf{28.07} & 31.61 & 29.49 & 38.35 & \textbf{41.54}\\
\textbf{Mr} & 15.12 & \textbf{18.89} & 23.96 & 17.46 & 36.12 & 27.54 & 39.98 & 31.46 & 41.89 & 41.10\\
\textbf{Od} & 11.11 & \textbf{15.09} & 9.06 & \textbf{15.52} & 14.20 & \textbf{23.76} & 12.78 & \textbf{24.37} & 28.81 & \textbf{36.39} \\
\textbf{Pa} & 15.60 & \textbf{20.54} & 28.43 & 19.88 & 21.96 & \textbf{30.54} & 32.04 & 29.03 & 43.37 & \textbf{45.10}\\
\textbf{Ta} & 19.96 & 18.22 & 22.45 & 17.42 & 19.74 & \textbf{27.54} & 28.04 & \textbf{28.06} & 39.64 & \textbf{40.29}\\
\textbf{Te} & 16.07 & \textbf{17.79} & 23.93 & 17.72 & 13.59 & \textbf{25.37}& 22.13 & \textbf{27.04} & 34.97 & \textbf{39.24}\\
\bottomrule
\end{tabular}%
}
\caption{Performance of both Direct Inference and Translate-Test Inference for various Large Language Models on the Zero-Shot Extractive Indic QA Benchmark. We report the average F1 scores across span-extraction datasets. Instances where Translate-Test outperforms Direct Inference are indicated in \textbf{bold}.}
\label{ref:ext}
\end{table*}

\begin{table*}[!h]
\centering
\resizebox{1.0\linewidth}{!}{%
\begin{tabular}{crrrrrrrrrrrrr} 
\toprule
\textbf{Languages} & \multicolumn{2}{c}{Bloom} & \multicolumn{2}{c}{Openhathi} & \multicolumn{2}{c}{Llama-3} & \multicolumn{2}{c}{Llama-3.1} & \multicolumn{2}{c}{Gemma} & \multicolumn{2}{c}{Gemma-2} \\
\cmidrule(lr){1-1} \cmidrule(lr){2-3} \cmidrule(lr){4-5} \cmidrule(lr){6-7} \cmidrule(lr){8-9} \cmidrule(lr){10-11} \cmidrule(lr){12-13}
\multicolumn{1}{l}{} & \begin{tabular}[c]{@{}c@{}}Direct\\ Inference\end{tabular} & \multicolumn{1}{c}{Translate} & \begin{tabular}[c]{@{}c@{}}Direct\\ Inference\end{tabular} & \multicolumn{1}{c}{Translate} & \begin{tabular}[c]{@{}c@{}}Direct\\ Inference\end{tabular} & \multicolumn{1}{c}{Translate} & \begin{tabular}[c]{@{}c@{}}Direct\\ Inference\end{tabular} & \multicolumn{1}{c}{Translate} & \begin{tabular}[c]{@{}c@{}}Direct\\ Inference\end{tabular} & \multicolumn{1}{c}{Translate} & \begin{tabular}[c]{@{}c@{}}Direct\\ Inference\end{tabular} & \multicolumn{1}{c}{Translate} \\
\midrule
\textbf{As} & 5.91 & \textbf{11.61} & 0.33 & \textbf{3.98} & 3.97 & \textbf{7.69} & 4.17 & \textbf{10.13} & 6.98 & \textbf{7.78} & 11.32 & \textbf{14.23} \\
\textbf{Bn} & 8.21 & \textbf{11.35} & 0.47 & \textbf{3.58} & 4.40 & \textbf{7.71} & 5.10 & \textbf{9.93} & 6.48 & \textbf{7.22} & 10.88 & \textbf{13.85} \\
\textbf{Gu} & 8.73 & \textbf{12.00} & 0.12 & \textbf{4.20} & 6.97 & \textbf{7.98} & 6.05 & \textbf{10.25} & 9.06 & 7.14 & 12.78 & \textbf{14.64} \\
\textbf{Hi} & 8.44 & \textbf{12.14} & 5.19 & 4.04 & 11.17 & 8.21 & 11.48 & 10.66 & 8.63 & 7.90 & 12.33 & \textbf{14.84} \\
\textbf{Kn} & 8.19 & \textbf{11.87} & 0.15 & \textbf{3.76} & 4.88 & \textbf{7.70} & 4.71 & \textbf{10.19} & 7.53 & \textbf{8.04} & 12.34 & \textbf{14.78} \\
\textbf{Ml} & 8.39 & \textbf{11.68} & 1.01 & \textbf{4.13} & 6.75 & \textbf{7.66} & 7.99 & \textbf{9.94} & 7.14 & \textbf{7.56} & 13.39 & \textbf{14.27} \\
\textbf{Mr} & 8.39 & \textbf{11.38} & 1.71 & \textbf{3.85} & 10.08 & 7.79 & 9.45 & \textbf{9.93} & 8.23 & 7.49 & 11.98 & \textbf{14.23} \\
\textbf{Od} & 6.84 & \textbf{11.58} & 0.00 & \textbf{3.68} & 7.01 & \textbf{7.61} & 7.88 & \textbf{9.93} & 3.38 & \textbf{6.77} & 11.70 & \textbf{14.12} \\
\textbf{Pa} & 7.91 & \textbf{11.90} & 0.15 & \textbf{4.18} & 6.10 & \textbf{8.09} & 5.81 & \textbf{10.28} & 9.72 & 7.84 & 12.96 & \textbf{14.90} \\
\textbf{Ta} & 9.52 & \textbf{12.66} & 0.47 & \textbf{4.34} & 3.72 & \textbf{7.95} & 6.17 & \textbf{10.46} & 7.74 & \textbf{7.94} & 13.67 & \textbf{15.38} \\
\textbf{Te} & 8.96 & \textbf{12.40} & 0.19 & \textbf{3.90} & 3.04 & \textbf{8.10} & 5.56 & \textbf{10.64} & 9.34 & 8.14 & 13.80 & \textbf{15.19} \\
\bottomrule
\end{tabular}%
}
\caption{Performance of both Direct Inference and translate test inference of various Large Language Models on Zero-Shot Abstractive INDIC QA BENCHMARK. We report the average numbers (Rouge-L) scores across the languages. Instances where Translate-Test outperforms Direct Inference are indicated in \textbf{bold}.}
\label{ref:gen1}
\end{table*}

\section{Experiments}

We conducted a series of experiments to evaluate the performance of existing LLMs, using NVIDIA A100 GPUs in both 40GB and 80GB variants for our computational needs. Our computational needs signify GPUs both for Translation and evaluation over the models. For inference we utilized VLLM \cite{kwon2023efficient} which is an open-source library that supports LLM inference efficiently. 

We evaluate the following LLMs on our benchmark: OpenHathi\footnote{\url{https://huggingface.co/sarvamai/OpenHathi-7B-Hi-v0.1-Base}} and its instruction-finetuned variant (IFV) known as Airavata \cite{gala2024airavata}, Bloom \cite{le2022bloom} and its IFV named Bloomz, Gemma\cite{team2024gemma}, and its instruction fine-tuned variant Gemma-IT. OpenHathi (7B parameter model), was created through continual pre-training on the LLaMA-2 model \cite{touvron2023llama}. Airavata \cite{gala2024airavata} (7B parameter model) is an instruction fine-tuned version of OpenHathi. Both OpenHathi and Airavata are specifically trained for Hindi. 

Gemma and Gemma-IT (7B parameter models)\footnote{\url{https://ai.google.dev/gemma/docs}} were released by Google. While these models are not specifically trained for Indian languages, they demonstrate multilingual capabilities. Additionally, Aya-8B \cite{aryabumi2024aya} is an instruction-tuned model specifically designed for multilingual applications. We also tested Aya-101, another instruction-tuned model based on MT5, which is available in a 13B size. Furthermore, we explored the LLaMA \{3, 3.1\} and LLaMA \{3, 3.1\} Instruct models\footnote{\url{https://ai.meta.com/blog/meta-llama-3/}}, which are 8B parameter models and part of the LLaMA family. LLaMA-3 has been trained on data from approximately 30 languages, excluding English. we also used Narvasa2.0 \footnote{\url{huggingface.co/Telugu-LLM-Labs/Indic-gemma-7b-finetuned-sft-Navarasa-2.0}} and Mistral NeMo \footnote{\url{https://mistral.ai/news/mistral-nemo/}} because of their strong multilingual performance. This diverse range of models enables a thorough evaluation of the strengths and performance of our benchmarks across different architectures and training methods. 

\begin{table*}[!h]
\centering
\resizebox{0.8\linewidth}{!}{%
\begin{tabular}{crrrrrrrrr} 
\toprule
\textbf{Languages} & \multicolumn{2}{c}{Bloom} & \multicolumn{2}{c}{Gemma} & \multicolumn{2}{c}{Llama-3} & \multicolumn{2}{c}{Openhathi} \\
\cmidrule(lr){1-1} \cmidrule(lr){2-3} \cmidrule(lr){4-5} \cmidrule(lr){6-7} \cmidrule(lr){8-9}
\multicolumn{1}{l}{} & \multicolumn{1}{c}{1 shot} & \multicolumn{1}{c}{3 shot} & \multicolumn{1}{c}{1 shot} & \multicolumn{1}{c}{3 shot} & \multicolumn{1}{c}{1 shot} & \multicolumn{1}{c}{3 shot} & \multicolumn{1}{c}{1 shot} & \multicolumn{1}{c}{3 shot}  \\
\midrule
\textbf{As} & 21.22 & 22.44 & 37.91 & 40.43 & 28.61 & 29.42 & 3.15 & 3.14\\
\textbf{Bn} &  19.82 & 26.90 & 42.64 & 37.85 & 32.71 & 24.95 & 6.86 & 6.56\\
\textbf{Gu} & 20.17  & 22.16 & 44.26 & 47.40 & 26.76 & 22.46 & 3.65 & 4.61\\
\textbf{Hi} & 34.08 & 36.96 & 54.95 & 56.95 & 54.79 & 57.11 & 10.27 & 22.1\\
\textbf{Kn} & 19.99 & 19.93 & 37.74 & 35.80 & 22.76 & 15.60 & 0.33 & 2.26\\
\textbf{Ml} & 25.82 & 23.84 & 42.08 & 31.10 & 28.94 & 22.60 & 6.59 & 5.50\\
\textbf{Mr} &  26.35 & 25.99  & 47.08 & 37.39 & 47.17 & 40.7 & 5.82 & 8.63\\
\textbf{Od} & 15.72  & 15.68 & 16.66 & 22.05 & 13.37 & 5.39 & 0.37 & 3.32\\
\textbf{Pa} & 23.18  & 23.98 & 42.18 & 47.17 & 31.03 & 30.62 & 2.00 & 3.52\\
\textbf{Ta} &  26.80 & 26.97 & 43.27 & 43.90 & 28.66 & 27.06 & 4.51 &  4.94\\
\textbf{Te} & 20.83 & 25.22 & 39.90 & 43.62 & 25.37 & 20.86 & 3.09 & 2.76\\
\bottomrule
\end{tabular}%
}
\caption{Performance of various Large Language Models on few-Shot Extractive INDIC QA BENCHMARK. We report the average (F1-score) across span-extraction datasets and question-answering datasets.}
\label{ref:resultFewShot}
\end{table*}

\subsection{Evaluation Metrics}

We chose widely used QA evaluation metrics for evaluating both extractive and abstractive Question Answering datasets: \\
    \noindent \textbf{1. F1 (macro-averaged) score:} This score represents the harmonic mean of precision and recall, calculating the average similarity between predicted and actual answers by comparing the sets of words or tokens in the predicted and ground truth sentences. \\
    \noindent \textbf{2. ROUGE(L):} A Recall-Oriented Understudy to measure how much of the words from the gold response are present in the generated response. This metric is commonly used for generative tasks such as summarization, and we have used it to evaluate abstractive QA tasks.

\subsection{Translate-Test Inferences}

In addition to direct inference, we conducted experiments on translate-test \ref{fig:res_summary} inferences involving the following steps:\\
    \noindent \textbf{1.} Use \textbf{IndicTrans2} to translate the source language input (context and question) to English.\\
    \noindent \textbf{2.} Prompt the LLM with the translated input and get the response.\\
    \noindent \textbf{3.} Back-translate the generated LLM's English response to the source language.

This method ensures the evaluation of the LLMs' performance across languages by utilizing translation systems to bridge the gap between languages during inference.

\section{Result and analysis}

\textbf{Comparing Base Model Performance and Effect of Few Shot:}
Table \ref{ref:result-main} shows the base LLMs' performance in the zero-shot setting. The Gemma model excels in extractive question-answering tasks, surpassing the Bloom and Llama-3 base models. However, Llama-3 outperforms Gemma in Hindi, Marathi, and Odia languages. Bloom surpasses all models in abstractive question-answering tasks and for all languages. Notably, base models generally perform poorly on abstractive question-answering datasets compared to extractive ones.

We also evaluate the effect of in-context examples as reported in Table~\ref{ref:resultFewShot}. As expected, using few-shot (1-shot and 3-shot) almost always improves over the zero-shot base model. However, we can spot some language-specific patterns where Bloom and Openhathi behave differently than Gemma and Llama-3. For example, for some languages such as \textit{Bn, Ml, Mr} Gemma and Llama-3 show a significant drop with the increase in few shot examples, however, Bloom and Openhathi retain or even improve. We believe this is correlated with the availability of language-specific corpus and their utilization in training these models.

\textbf{Effect of Instruction Finetuning:}
Table \ref{ref:result-main} shows the performance of instruction-finetuned models in our study. Instruction finetuning generally improves abstractive QA tasks across all models, but its impact on extractive QA varies. For instance, Gemma and Llama-3 perform better than Bloom and OpenHathi in their base models, but their instruction-finetuned variants do not show significant improvement. This is because these models were primarily instruction-finetuned on non-Indic languages, which compromises their generic multi-lingual ability during task-specific finetuning, leading to lower results.

On the other hand, OpenHathi was specifically trained on the Hindi language and so is its instruction finetuning variant Airavata. As a result, the performance of OpenHathi is significantly poor in all languages. Airavata benefits from further instruction finetuning on Hindi data and improves over OpenHathi for Hindi language but suffers poorly for other Indian languages. Bloomz produces the highest jump compared to Bloom and we hypothesize this is because a good portion of evaluation benchmark coming from generic-domain such as Wikipedia data has been seen by Bloomz during its training and instruction finetuning, making it a good choice for applications which aims to use common world knowledge.

\textbf{Extractive vs Abstractive Tasks:}
While it is clear that instruction finetuning helps more in abstractive QA tasks, both Table~\ref{ref:subset} and Table~\ref{ref:result-main} show a positive correlation between the scores for extractive tasks and abstractive tasks across languages. This is almost true for all the base and instruct variants of the models except Gemma, where Gemma instruct improves the abstractive QA score but deteriorates in the extractive QA task. Careful analysis shows that abstractive task metrics change moderately between base models and their instruction finetuned variants. This is expected because abstractive  metrics such as Rouge-L are more heuristic-driven in nature and designed to ignore small variations, natural in non-deterministic text generation, unlike extractive metrics such as F1. Thus abstractive metrics deviate on a smaller scale than extractive metrics. However, the positive correlation between both task metrics across models clearly establishes that the factors affecting the overall performance of the models show similar signs for both extractive and abstractive tasks and hence improving one will likely improve the other as well.

\textbf{On How to Choose a Model:}
Going by the results so far, one would pick BloomZ if the application needs only common world knowledge and needs a model which does well out-of-the-box. If there is a use-case for which we have adequate Indic language finetuning data, it might be good to build over the world knowledge acquired by Gemma and Llama-3 and do instruction finetuning on Indic languages to make it better suitable for abstractive QA tasks. If we are very specific about a certain niche domain in only the Hindi language, where common world knowledge is not a pre-requisite, Airavata can be a good candidate given its focus on Hindi-based training and improvements in both extractive and abstractive tasks with instruction finetuning. However, the Aya models are particularly well-suited if we are specifically seeking high-quality instruction-tuned models for Indian languages.

\textbf{Translate Test is an Effective Alternative to Source Language:}
Translation-based approaches are often more effective than direct generation in source languages using multilingual models. For languages like Punjabi, Gujarati, and Oriya, the translation-based approach outperforms the multilingual model. This comparison divides languages into two categories: (1) those where multilingual models perform better and (2) those where translation-based approaches are superior.

Mid-resource languages like Hindi and Bengali benefit more from multilingual models, while low-resource languages like Oriya and Punjabi perform better with translation-based approaches. This is because multilingual models struggle with insufficient language-specific data, leading to poor performance. In contrast, translation-based approaches leverage high-resource languages for reasoning and generation, making it easier to learn the translation task from multilingual data. 

\section{Conclusion}

In this paper, we present a benchmark for evaluating the grounded Question-Answering (QA) capabilities of Large Language Models (LLMs) on both extractive and abstractive tasks. Our findings reveal that instruction-tuning with target language data significantly enhances QA performance, while the Translate-Test technique yields better results for low-resource languages. In contrast, high-resource languages benefit more from source language inference due to larger training datasets.

Despite advances in multilingual training, LLMs struggle with transfer learning to low-resource languages. Integrating effective translation systems into the QA pipeline is crucial for improving performance in these contexts. By releasing this benchmark, we aim to promote further research into the QA capabilities of LLMs across various languages, particularly for those that are underrepresented.

\section{Limitation}
Our research aims to provide a challenging and comprehensive benchmark for evaluating LLMs on the Hindi QA task, but it does face several limitations.

\textbf{(1)} The availability of high-quality datasets for Hindi is limited. Despite our best efforts to curate the benchmark from various sources, there might still be an inherent bias introduced during the data collection and translation process. Additionally, although we conducted quality checks, there may be subjective interpretability issues with the translated datasets. \textbf{(2)} while we attempted to diversify across various domains, the benchmark may not depict the true performance in a completely unseen domain. 

Despite the strong performance of the Translate Test technique, particularly for low-resource languages, it has several limitations. One significant drawback is the potential for cascading errors. Translation errors occurring early in the pipeline can propagate through subsequent stages, adversely affecting the final output. This issue is typical in models that rely on sequential processing, where initial inaccuracies can compound over time. 
Moreover, while the Translate Test approach currently shows promising results, the ideal solution is the development of robust multilingual models that can handle and mitigate cascading errors effectively. Such models should be capable of generalizing well across various languages without relying heavily on error-prone translation processes. Our analysis focused on base models rather than instruction-tuned models. Previous research has shown that multilingual instruction-tuned  A well-optimized multilingual instruction-tuned model could potentially address these limitations. 

However, the challenge remains in developing effective instruction-tuning data for low-resource languages, whether through translation or other methods. This underscores the need for continued research to enhance instruction-tuning strategies in multilingual settings.
\section*{Acknowledgments}
We thank the anonymous reviewers for their constructive feedback. This work is supported by and is part of BharatGen (\url{https://bharatgen.tech/}), an Indian Government-funded initiative focused on developing multimodal large language models for Indian languages.

\bibliography{acl_latex}

\begin{thebibliography}{31}
\providecommand{\natexlab}[1]{#1}

\bibitem[{Artetxe et~al.(2019)Artetxe, Ruder, and Yogatama}]{artetxe2019cross}
Mikel Artetxe, Sebastian Ruder, and Dani Yogatama. 2019.
\newblock On the cross-lingual transferability of monolingual representations.
\newblock \emph{arXiv preprint arXiv:1910.11856}.

\bibitem[{Aryabumi et~al.(2024)Aryabumi, Dang, Talupuru, Dash, Cairuz, Lin, Venkitesh, Smith, Campos, Tan, Marchisio, Bartolo, Ruder, Locatelli, Kreutzer, Frosst, Gomez, Blunsom, Fadaee, Üstün, and Hooker}]{aryabumi2024aya}
Viraat Aryabumi, John Dang, Dwarak Talupuru, Saurabh Dash, David Cairuz, Hangyu Lin, Bharat Venkitesh, Madeline Smith, Jon~Ander Campos, Yi~Chern Tan, Kelly Marchisio, Max Bartolo, Sebastian Ruder, Acyr Locatelli, Julia Kreutzer, Nick Frosst, Aidan Gomez, Phil Blunsom, Marzieh Fadaee, Ahmet Üstün, and Sara Hooker. 2024.
\newblock \href {https://arxiv.org/abs/2405.15032} {Aya 23: Open weight releases to further multilingual progress}.
\newblock \emph{Preprint}, arXiv:2405.15032.

\bibitem[{Asai et~al.(2020)Asai, Kasai, Clark, Lee, Choi, and Hajishirzi}]{asai2020xor}
Akari Asai, Jungo Kasai, Jonathan~H Clark, Kenton Lee, Eunsol Choi, and Hannaneh Hajishirzi. 2020.
\newblock Xor qa: Cross-lingual open-retrieval question answering.
\newblock \emph{arXiv preprint arXiv:2010.11856}.

\bibitem[{Bonifacio et~al.(2021)Bonifacio, Campiotti, de~Alencar~Lotufo, and Nogueira}]{DBLP:journals/corr/abs-2108-13897}
Luiz Bonifacio, Israel Campiotti, Roberto de~Alencar~Lotufo, and Rodrigo~Frassetto Nogueira. 2021.
\newblock \href {https://arxiv.org/abs/2108.13897} {mmarco: {A} multilingual version of {MS} {MARCO} passage ranking dataset}.
\newblock \emph{CoRR}, abs/2108.13897.

\bibitem[{Chen et~al.(2024)Chen, Xiao, Zhang, Luo, Lian, and Liu}]{bge-m3}
Jianlv Chen, Shitao Xiao, Peitian Zhang, Kun Luo, Defu Lian, and Zheng Liu. 2024.
\newblock \href {https://arxiv.org/abs/2402.03216} {Bge m3-embedding: Multi-lingual, multi-functionality, multi-granularity text embeddings through self-knowledge distillation}.
\newblock \emph{Preprint}, arXiv:2402.03216.

\bibitem[{Conneau et~al.(2019)Conneau, Khandelwal, Goyal, Chaudhary, Wenzek, Guzm{\'a}n, Grave, Ott, Zettlemoyer, and Stoyanov}]{conneau2019unsupervised}
Alexis Conneau, Kartikay Khandelwal, Naman Goyal, Vishrav Chaudhary, Guillaume Wenzek, Francisco Guzm{\'a}n, Edouard Grave, Myle Ott, Luke Zettlemoyer, and Veselin Stoyanov. 2019.
\newblock Unsupervised cross-lingual representation learning at scale.
\newblock \emph{arXiv preprint arXiv:1911.02116}.

\bibitem[{Doddapaneni et~al.(2022)Doddapaneni, Aralikatte, Ramesh, Goyal, Khapra, Kunchukuttan, and Kumar}]{Doddapaneni2022towards}
Sumanth Doddapaneni, Rahul Aralikatte, Gowtham Ramesh, Shreyansh Goyal, Mitesh~M. Khapra, Anoop Kunchukuttan, and Pratyush Kumar. 2022.
\newblock Towards leaving no indic language behind: Building monolingual corpora, benchmark and models for indic languages.
\newblock \emph{ArXiv}, abs/2212.05409.

\bibitem[{Dzendzik et~al.(2021)Dzendzik, Vogel, and Foster}]{dzendzik2021english}
Daria Dzendzik, Carl Vogel, and Jennifer Foster. 2021.
\newblock English machine reading comprehension datasets: A survey.
\newblock \emph{arXiv preprint arXiv:2101.10421}.

\bibitem[{Etxaniz et~al.(2023)Etxaniz, Azkune, Soroa, de~Lacalle, and Artetxe}]{etxaniz2023multilingual}
Julen Etxaniz, Gorka Azkune, Aitor Soroa, Oier~Lopez de~Lacalle, and Mikel Artetxe. 2023.
\newblock Do multilingual language models think better in english?
\newblock \emph{arXiv preprint arXiv:2308.01223}.

\bibitem[{Gala et~al.(2023)Gala, Chitale, AK, Gumma, Doddapaneni, Kumar, Nawale, Sujatha, Puduppully, Raghavan et~al.}]{gala2023indictrans2}
Jay Gala, Pranjal~A Chitale, Raghavan AK, Varun Gumma, Sumanth Doddapaneni, Aswanth Kumar, Janki Nawale, Anupama Sujatha, Ratish Puduppully, Vivek Raghavan, et~al. 2023.
\newblock Indictrans2: Towards high-quality and accessible machine translation models for all 22 scheduled indian languages.
\newblock \emph{arXiv preprint arXiv:2305.16307}.

\bibitem[{Gala et~al.(2024)Gala, Jayakumar, Husain, M, Khan, Kanojia, Puduppully, Khapra, Dabre, Murthy, and Kunchukuttan}]{gala2024airavata}
Jay Gala, Thanmay Jayakumar, Jaavid~Aktar Husain, Aswanth~Kumar M, Mohammed Safi Ur~Rahman Khan, Diptesh Kanojia, Ratish Puduppully, Mitesh~M. Khapra, Raj Dabre, Rudra Murthy, and Anoop Kunchukuttan. 2024.
\newblock Airavata: Introducing hindi instruction-tuned llm.
\newblock \emph{arXiv preprint arXiv: 2401.15006}.

\bibitem[{Intrator et~al.(2024)Intrator, Halfon, Goldenberg, Tsarfaty, Eyal, Rivlin, Matias, and Aizenberg}]{intrator2024breaking}
Yotam Intrator, Matan Halfon, Roman Goldenberg, Reut Tsarfaty, Matan Eyal, Ehud Rivlin, Yossi Matias, and Natalia Aizenberg. 2024.
\newblock Breaking the language barrier: Can direct inference outperform pre-translation in multilingual llm applications?
\newblock \emph{arXiv preprint arXiv:2403.04792}.

\bibitem[{Kolhatkar and Verma(2023)}]{kolhatkar2023indic}
Dhruv Kolhatkar and Devika Verma. 2023.
\newblock Indic language question answering: A survey.
\newblock In \emph{2023 Third International Conference on Artificial Intelligence and Smart Energy (ICAIS)}, pages 697--703. IEEE.

\bibitem[{Kwon et~al.(2023)Kwon, Li, Zhuang, Sheng, Zheng, Yu, Gonzalez, Zhang, and Stoica}]{kwon2023efficient}
Woosuk Kwon, Zhuohan Li, Siyuan Zhuang, Ying Sheng, Lianmin Zheng, Cody~Hao Yu, Joseph Gonzalez, Hao Zhang, and Ion Stoica. 2023.
\newblock Efficient memory management for large language model serving with pagedattention.
\newblock In \emph{Proceedings of the 29th Symposium on Operating Systems Principles}, pages 611--626.

\bibitem[{Le~Scao et~al.(2022)Le~Scao, Fan, Akiki, Pavlick, Ili{\'c}, Hesslow, Castagn{\'e}, Luccioni, Yvon, Gall{\'e} et~al.}]{le2022bloom}
Teven Le~Scao, Angela Fan, Christopher Akiki, Ellie Pavlick, Suzana Ili{\'c}, Daniel Hesslow, Roman Castagn{\'e}, Alexandra~Sasha Luccioni, Fran{\c{c}}ois Yvon, Matthias Gall{\'e}, et~al. 2022.
\newblock Bloom: A 176b-parameter open-access multilingual language model.

\bibitem[{Lee et~al.(2019)Lee, Chang, and Toutanova}]{lee-etal-2019-latent}
Kenton Lee, Ming-Wei Chang, and Kristina Toutanova. 2019.
\newblock \href {https://doi.org/10.18653/v1/P19-1612} {Latent retrieval for weakly supervised open domain question answering}.
\newblock In \emph{Proceedings of the 57th Annual Meeting of the Association for Computational Linguistics}, pages 6086--6096, Florence, Italy. Association for Computational Linguistics.

\bibitem[{Lewis et~al.(2019)Lewis, O\u{g}uz, Rinott, Riedel, and Schwenk}]{lewis2019mlqa}
Patrick Lewis, Barlas O\u{g}uz, Ruty Rinott, Sebastian Riedel, and Holger Schwenk. 2019.
\newblock Mlqa: Evaluating cross-lingual extractive question answering.
\newblock \emph{arXiv preprint arXiv:1910.07475}.

\bibitem[{Liu(2022)}]{Liu_LlamaIndex_2022}
Jerry Liu. 2022.
\newblock \href {https://doi.org/10.5281/zenodo.1234} {{LlamaIndex}}.

\bibitem[{Penedo et~al.(2023)Penedo, Malartic, Hesslow, Cojocaru, Cappelli, Alobeidli, Pannier, Almazrouei, and Launay}]{penedo2023refinedwebdatasetfalconllm}
Guilherme Penedo, Quentin Malartic, Daniel Hesslow, Ruxandra Cojocaru, Alessandro Cappelli, Hamza Alobeidli, Baptiste Pannier, Ebtesam Almazrouei, and Julien Launay. 2023.
\newblock \href {https://arxiv.org/abs/2306.01116} {The refinedweb dataset for falcon llm: Outperforming curated corpora with web data, and web data only}.
\newblock \emph{Preprint}, arXiv:2306.01116.

\bibitem[{Popovi{\'c}(2015)}]{popovic2015chrf}
Maja Popovi{\'c}. 2015.
\newblock chrf: character n-gram f-score for automatic mt evaluation.
\newblock In \emph{Proceedings of the tenth workshop on statistical machine translation}, pages 392--395.

\bibitem[{Rahman~Khan et~al.(2024)Rahman~Khan, Mehta, Sankar, Kumaravelan, Doddapaneni, Balan~G, Jain, Kunchukuttan, Kumar, Dabre et~al.}]{rahman2024indicllmsuite}
Mohammed Safi~Ur Rahman~Khan, Priyam Mehta, Ananth Sankar, Umashankar Kumaravelan, Sumanth Doddapaneni, Varun Balan~G, Sparsh Jain, Anoop Kunchukuttan, Pratyush Kumar, Raj Dabre, et~al. 2024.
\newblock Indicllmsuite: A blueprint for creating pre-training and fine-tuning datasets for indian languages.
\newblock \emph{arXiv e-prints}, pages arXiv--2403.

\bibitem[{Rajpurkar et~al.(2018)Rajpurkar, Jia, and Liang}]{rajpurkar2018know}
Pranav Rajpurkar, Robin Jia, and Percy Liang. 2018.
\newblock Know what you don’t know: Unanswerable questions for squad.
\newblock In \emph{Proceedings of the 56th Annual Meeting of the Association for Computational Linguistics (Volume 2: Short Papers)}. Association for Computational Linguistics.

\bibitem[{Rajpurkar et~al.(2016)Rajpurkar, Zhang, Lopyrev, and Liang}]{rajpurkar2016squad}
Pranav Rajpurkar, Jian Zhang, Konstantin Lopyrev, and Percy Liang. 2016.
\newblock Squad: 100,000+ questions for machine comprehension of text.
\newblock \emph{arXiv preprint arXiv:1606.05250}.

\bibitem[{Reid and Artetxe(2022)}]{reid2022role}
Machel Reid and Mikel Artetxe. 2022.
\newblock On the role of parallel data in cross-lingual transfer learning.
\newblock \emph{arXiv preprint arXiv:2212.10173}.

\bibitem[{Singh et~al.(2024)Singh, Gupta, Bharadwaj, Tewari, and Talukdar}]{singh2024indicgenbench}
Harman Singh, Nitish Gupta, Shikhar Bharadwaj, Dinesh Tewari, and Partha Talukdar. 2024.
\newblock Indicgenbench: A multilingual benchmark to evaluate generation capabilities of llms on indic languages.
\newblock \emph{arXiv preprint arXiv:2404.16816}.

\bibitem[{Team et~al.(2023)Team, Anil, Borgeaud, Wu, Alayrac, Yu, Soricut, Schalkwyk, Dai, Hauth et~al.}]{team2023gemini}
Gemini Team, Rohan Anil, Sebastian Borgeaud, Yonghui Wu, Jean-Baptiste Alayrac, Jiahui Yu, Radu Soricut, Johan Schalkwyk, Andrew~M Dai, Anja Hauth, et~al. 2023.
\newblock Gemini: a family of highly capable multimodal models.
\newblock \emph{arXiv preprint arXiv:2312.11805}.

\bibitem[{Team et~al.(2024)Team, Mesnard, Hardin, Dadashi, Bhupatiraju, Pathak, Sifre, Rivi{\`e}re, Kale, Love et~al.}]{team2024gemma}
Gemma Team, Thomas Mesnard, Cassidy Hardin, Robert Dadashi, Surya Bhupatiraju, Shreya Pathak, Laurent Sifre, Morgane Rivi{\`e}re, Mihir~Sanjay Kale, Juliette Love, et~al. 2024.
\newblock Gemma: Open models based on gemini research and technology.
\newblock \emph{arXiv preprint arXiv:2403.08295}.

\bibitem[{Thakur et~al.(2021)Thakur, Reimers, R{\"u}ckl{\'e}, Srivastava, and Gurevych}]{thakur2021beir}
Nandan Thakur, Nils Reimers, Andreas R{\"u}ckl{\'e}, Abhishek Srivastava, and Iryna Gurevych. 2021.
\newblock \href {https://openreview.net/forum?id=wCu6T5xFjeJ} {{BEIR}: A heterogeneous benchmark for zero-shot evaluation of information retrieval models}.
\newblock In \emph{Thirty-fifth Conference on Neural Information Processing Systems Datasets and Benchmarks Track (Round 2)}.

\bibitem[{Thirumala and Ferracane(2022)}]{thirumala2022extractive}
Adhitya Thirumala and Elisa Ferracane. 2022.
\newblock Extractive question answering on queries in hindi and tamil.
\newblock \emph{arXiv preprint arXiv:2210.06356}.

\bibitem[{Touvron et~al.(2023)Touvron, Martin, Stone, Albert, Almahairi, Babaei, Bashlykov, Batra, Bhargava, Bhosale, Bikel, Blecher, Ferrer, Chen, Cucurull, Esiobu, Fernandes, Fu, Fu, Fuller, Gao, Goswami, Goyal, Hartshorn, Hosseini, Hou, Inan, Kardas, Kerkez, Khabsa, Kloumann, Korenev, Koura, Lachaux, Lavril, Lee, Liskovich, Lu, Mao, Martinet, Mihaylov, Mishra, Molybog, Nie, Poulton, Reizenstein, Rungta, Saladi, Schelten, Silva, Smith, Subramanian, Tan, Tang, Taylor, Williams, Kuan, Xu, Yan, Zarov, Zhang, Fan, Kambadur, Narang, Rodriguez, Stojnic, Edunov, and Scialom}]{touvron2023llama}
Hugo Touvron, Louis Martin, Kevin Stone, Peter Albert, Amjad Almahairi, Yasmine Babaei, Nikolay Bashlykov, Soumya Batra, Prajjwal Bhargava, Shruti Bhosale, Dan Bikel, Lukas Blecher, Cristian~Canton Ferrer, Moya Chen, Guillem Cucurull, David Esiobu, Jude Fernandes, Jeremy Fu, Wenyin Fu, Brian Fuller, Cynthia Gao, Vedanuj Goswami, Naman Goyal, Anthony Hartshorn, Saghar Hosseini, Rui Hou, Hakan Inan, Marcin Kardas, Viktor Kerkez, Madian Khabsa, Isabel Kloumann, Artem Korenev, Punit~Singh Koura, Marie-Anne Lachaux, Thibaut Lavril, Jenya Lee, Diana Liskovich, Yinghai Lu, Yuning Mao, Xavier Martinet, Todor Mihaylov, Pushkar Mishra, Igor Molybog, Yixin Nie, Andrew Poulton, Jeremy Reizenstein, Rashi Rungta, Kalyan Saladi, Alan Schelten, Ruan Silva, Eric~Michael Smith, Ranjan Subramanian, Xiaoqing~Ellen Tan, Binh Tang, Ross Taylor, Adina Williams, Jian~Xiang Kuan, Puxin Xu, Zheng Yan, Iliyan Zarov, Yuchen Zhang, Angela Fan, Melanie Kambadur, Sharan Narang, Aurelien Rodriguez, Robert Stojnic, Sergey Edunov, and Thomas
  Scialom. 2023.
\newblock \href {https://arxiv.org/abs/2307.09288} {Llama 2: Open foundation and fine-tuned chat models}.
\newblock \emph{Preprint}, arXiv:2307.09288.

\bibitem[{Yang et~al.(2018)Yang, Qi, Zhang, Bengio, Cohen, Salakhutdinov, and Manning}]{yang2018hotpotqa}
Zhilin Yang, Peng Qi, Saizheng Zhang, Yoshua Bengio, William~W Cohen, Ruslan Salakhutdinov, and Christopher~D Manning. 2018.
\newblock Hotpotqa: A dataset for diverse, explainable multi-hop question answering.
\newblock \emph{arXiv preprint arXiv:1809.09600}.

\end{thebibliography}
\appendix
\section{Appendix} \label{appendix}
\subsection{Synthetic Data Creation} \label{syn_data}
As discussed in \ref{dataset_section}, in addition to the translation dataset from high-resource to low-resource languages, we also generated data using the Gemini model \cite{team2023gemini}. We chose this model due to its strong multilingual performance and considerable parameter count. To start, we sampled paragraphs from various sources, including Wikipedia, storybooks, Indian news articles, and a selection of books, aiming for domain diversity while incorporating Indian cultural nuances.
We developed our benchmark in Hindi and selected a subset of the generated data for human verification. The verification process focused on several parameters, such as the fluency of the generated questions and answers, and whether the questions were directly derived from the paragraphs. Given that large language models can hallucinate even with clear instructions annotators were asked to score each instance on a scale of 1 to 5, where 1 indicated irrelevance and 5 indicated exact relevance. We also evaluated whether the answers were generated from the text or relied on external knowledge.

Once verified, we translated these data instances into other Indian languages. The synthetic data primarily consisted of texts from Ramayana textbooks, recent Indian news articles, cultural articles from Wikipedia, and classic storybooks, Dharmpal books. Most of the data was extracted from books using Optical Character Recognition (OCR), followed by a thorough cleaning process to compile useful passages.

\subsection{Analysis}
\textbf{Indic Languages are Medium or Low Resource:}
Before diving into the results, it's important to consider the resource availability in Indic languages. This influences the best strategy for QA tasks in these languages. Figure~\ref{fig:sangraha_stas} shows the approximate number of tokens available for each language from sources like Wikipedia, websites, and PDFs. Compared to English, which has trillions of tokens, Indic languages have far fewer resources. Languages like Hindi and Bengali are medium-resource, while most others are low-resource, with some like Oriya and Punjabi being very low-resource. Although the statistics in Figure~\ref{fig:sangraha_stas} are estimates and change over time and across different models, classifying languages into high, medium, and low-resource groups is important for our analysis.

Tables \ref{ref:ext} and \ref{ref:gen} present important observations, which we outline below:

\textbf{Extractive vs Abstractive Shows Same Pattern:}
The patterns shown by both the extractive and abstractive dataset results in Tables \ref{ref:ext} and \ref{ref:gen} are consistent. This is due to the impact of the availability or lack of language-specific training data, and the quality of translation remaining the same for both types of QA tasks. Therefore, the insights and patterns observed in Table \ref{ref:ext} also apply to Table \ref{ref:gen}.

\textbf{Gemma2-Base performed best on our experiments:}
There is noticeable variation in the performance of different LLMs across languages. This can be attributed to the varying amounts of language-specific training data and the different sizes of the models. Despite this variation, the observation that translation-based methods work better for low-resource languages generally holds true across all LLMs.

Among the LLMs, Gemma2 is the clear winner across most languages and tasks. Llama-3 and Gemma perform similarly and slightly better than Bloom. The performance ranking of LLMs tends to follow their release order, suggesting that more training data improves their multilingual capabilities. Openhathi performs relatively poorly compared to other models, likely because it has only been trained on Hindi data and has limited exposure to other languages.

Additionally, Aya-101, Narvasa 2.0, and Mistral NeMo are superior instruction-tuned models. This can be attributed to their fine-tuning on Indic data, whereas other models perform well in their base variants but struggle during instruction fine-tuning. This difficulty is likely due to catastrophic forgetting, as the fine-tuning data for these models contains significantly less Indic data compared to their pre-training datasets.
Aya-101 is fine-tuned over the base model mT5, which is an encoder-decoder model (non-LLM). Although it is 13B parameter model which is greater than other in terms of parameter count, then to its performance is way higher than other multilinguual instruction tunned models [\ref{ref:subset}].

\begin{table*}[!h]
\centering
\resizebox{1.0\linewidth}{!}{%
\begin{tabular}{crrrrrrrrrrr} 
\toprule
\textbf{Languages} & \multicolumn{2}{c}{Llam3.1 Instruct} & \multicolumn{2}{c}{Gemma2-it} & \multicolumn{2}{c}{Navarasa-2.0} & \multicolumn{2}{c}{Mistral-nemo} & \multicolumn{2}{c}{Aya-101} \\
\cmidrule(lr){1-1} \cmidrule(lr){2-3} \cmidrule(lr){4-5} \cmidrule(lr){6-7} \cmidrule(lr){8-9} \cmidrule(lr){10-11}
\multicolumn{1}{l}{} & \multicolumn{1}{c}{Ext} & \multicolumn{1}{c}{Gen} & \multicolumn{1}{c}{Ext} & \multicolumn{1}{c}{Gen} & \multicolumn{1}{c}{Ext} & \multicolumn{1}{c}{Gen} & \multicolumn{1}{c}{Ext} & \multicolumn{1}{c}{Gen} & \multicolumn{1}{c}{Ext} & \multicolumn{1}{c}{Gen} \\

\midrule
\textbf{As} & 13.82  & 3.33 & 3.43 & 1.32 & 21.59 & 3.27 & 21.29 & 0.93 & 43.38 & 3.15 \\
\textbf{Bn} & 15.96 & 3.69 & 5.33 & 1.98 & 19.14 & 3.59 & 25.35 & 1.57 & 57.64 & 3.79\\
\textbf{Gu} & 12.73 & 5.80 & 6.19 & 3.37 & 30.77 & 6.01 & 24.08 & 2.61 & 61.42 & 5.84\\
\textbf{Hi} & 25.31 & 3.93 & 10.63 & 2.64 & 31.31 & 5.62 & 46.76 & 2.03 & 70.45 & 2.43\\
\textbf{Kn} & 6.37 & 4.46 & 3.35 & 1.85 & 24.55 & 4.99 & 21.31 & 1.24 & 55.21 & 4.23\\
\textbf{Ml} & 14.56 & 5.88 & 4.42 & 2.10 & 28.40 & 5.21 & 31.61 & 2.13 & 55.52 & 5.25\\
\textbf{Mr} & 24.37 & 4.68 & 8.48 & 2.65 & 22.90 & 2.78 & 39.98 & 1.28 & 58.98 & 3.34\\
\textbf{Od} & 14.56 & 5.04 & 2.99 & 0.95 & 20.75 & 3.12 & 0.26 & 0.60 & 48.82 & 3.23 \\
\textbf{Pa} & 10.25 & 4.21 & 3.59 & 1.73 & 25.84 & 5.88 & 32.04 & 2.42 & 64.19 & 6.01\\
\textbf{Ta} & 13.95 & 6.23 & 5.75 & 2.97 & 18.67 & 6.57 & 28.04 & 2.40 & 58.27 & 7.07\\
\textbf{Te} & 10.33 & 5.32 & 4.09 & 2.19 & 21.94 & 5.89 & 22.13 & 1.66 & 56.60 & 7.29\\
\bottomrule
\end{tabular}%
}
\caption{Performance of various Large Language Models on Zero-Shot on \textbf{Subset} of INDIC QA BENCHMARK (MLQA, NQ open, Synthetic data). We report the average numbers (F1) across span-extraction datasets and (Rouge-L)abstractive question-answering datasets.}
\label{ref:subset}
\end{table*}


\begin{table*}[!htb]
\centering
\resizebox{0.9\linewidth}{!}{%
\begin{tabular}{crrrrrrrrrrrrr} 
\toprule
\textbf{Languages} & \multicolumn{2}{c}{Bloomz} & \multicolumn{2}{c}{Gemma-Instruct} & \multicolumn{2}{c}{Llama-3-Instruct} & \multicolumn{2}{c}{Airavata} & \multicolumn{2}{c}{Aya-8-Instruct} \\
\cmidrule(lr){1-1} \cmidrule(lr){2-3} \cmidrule(lr){4-5} \cmidrule(lr){6-7} \cmidrule(lr){8-9} \cmidrule(lr){10-11} 
\multicolumn{1}{l}{} & \multicolumn{1}{c}{Ext} & \multicolumn{1}{c}{Gen} & \multicolumn{1}{c}{Ext} & \multicolumn{1}{c}{Gen} & \multicolumn{1}{c}{Ext} & \multicolumn{1}{c}{Gen} & \multicolumn{1}{c}{Ext} & \multicolumn{1}{c}{Gen} & \multicolumn{1}{c}{Ext} & \multicolumn{1}{c}{Gen} \\
\midrule
\textbf{As} & \textbf{38.69} & \textbf{7.36} & 11.32 & 9.97 & 15.22 & 3.42 & 3.70 & 3.41 &  19.39 & 7.19\\
\textbf{Bn} & \textbf{44.75} & \textbf{8.27} & 13.03 & 11.66 & 17.89 & 3.90 & 6.58 & 4.38 &  29.79 & \textbf{8.27}\\
\textbf{Gu} & \textbf{49.06} & \textbf{9.24} & 7.61 & 7.74 & 12.50 & 2.90 & 5.15 & 2.87 & 26.48 & 5.82\\
\textbf{Hi} & \textbf{62.88} & \textbf{9.06} & 18.54 & 7.61 & 14.95 & 7.21 & 44.35 & 7.21 & 55.17 & 8.02\\
\textbf{Kn} & \textbf{40.10} & 8.27 & 11.94 & 8.80 & 15.88 & 3.13 & 1.39 & 1.07 & 16.22 & \textbf{8.62}\\
\textbf{Ml} & \textbf{43.84} & \textbf{8.55} & 9.33 & 5.28 & 13.67 & 2.32 & 7.55 & 6.80 & 30.35 & \textbf{8.55}\\
\textbf{Mr} & \textbf{50.03} & 8.09 & 14.15 & 10.95 & 13.84 & 1.87 & 14.15 & 5.91 &  35.83 & \textbf{9.78}\\
\textbf{Od} & \textbf{37.10} & \textbf{8.78} & 1.61 & 0.72 & 5.59 & 1.74 &  1.78 & 1.27 & 15.77 & 5.79\\
\textbf{Pa} & \textbf{53.37} & 9.11 & 10.78 & 8.23 & 18.07 & 4.72 & 3.53 & 2.87 & 11.81 & \textbf{9.79}\\
\textbf{Ta} & \textbf{46.43} & \textbf{10.33} & 16.31 & 10.42 & 17.98 & 3.90 & 7.66 & 5.01 & 36.46 & 10.26\\
\textbf{Te} & \textbf{44.42} & \textbf{8.98} & 10.85 & 7.53 & 15.04 & 4.13 & 4.43 & 3.48 & 18.82 & 7.61\\
\midrule
\textbf{Average} & 46.42 & 8.73 & 11.41 & 8.08 & 14.60 & 3.57 & 9.12 & 4.03 & 26.92 & 8.15\\
\bottomrule
\end{tabular}%
}
\caption{Performance of various Large Language Models on Zero-Shot INDIC QA BENCHMARK. We report the average numbers (F1) across span-extraction datasets and (Rouge-L)abstractive question-answering datasets.}
\label{ref:result-main}
\end{table*}

\begin{table*}[!h]
\centering
\resizebox{1.0\linewidth}{!}{%
\begin{tabular}{crrrrrrrrrrr} 
\toprule
\textbf{Languages} & \multicolumn{2}{c}{Bloom} & \multicolumn{2}{c}{Gemma} & \multicolumn{2}{c}{Llama-3} & \multicolumn{2}{c}{Openhathi} & \multicolumn{2}{c}{Gemma-2} \\
\cmidrule(lr){1-1} \cmidrule(lr){2-3} \cmidrule(lr){4-5} \cmidrule(lr){6-7} \cmidrule(lr){8-9} \cmidrule(lr){10-11}
\multicolumn{1}{l}{} & \begin{tabular}[c]{@{}c@{}}Direct\\ Inference\end{tabular} & \multicolumn{1}{c}{Translate} & \begin{tabular}[c]{@{}c@{}}Direct\\ Inference\end{tabular} & \multicolumn{1}{c}{Translate} & \begin{tabular}[c]{@{}c@{}}Direct\\ Inference\end{tabular} & \multicolumn{1}{c}{Translate} & \begin{tabular}[c]{@{}c@{}}Direct\\ Inference\end{tabular} & \multicolumn{1}{c}{Translate} & \begin{tabular}[c]{@{}c@{}}Direct\\ Inference\end{tabular} & \multicolumn{1}{c}{Translate} \\
\midrule
\textbf{As} & 13.86  & \textbf{16.19} & 21.70 & 16.08 & 19.23 & \textbf{24.14} & 1.14 & \textbf{5.38} & 33.01 & \textbf{35.60} \\
\textbf{Bn} & 17.84 & 16.07 & 21.15 & 16.69 & 19.14 & \textbf{23.96} & 1.49 & \textbf{5.46} & 38.33 & 35.11\\
\textbf{Gu} & 13.27 & \textbf{18.59} & 24.90 & \textbf{25.12} & 20.27 & \textbf{29.55} & 4.07 & \textbf{7.15} & 40.10 & \textbf{44.86}\\
\textbf{Hi} & 21.69 & 19.29 & 34.37 & 19.72 & 41.53 & 30.02 & 11.83 & 8.28 & 44.46 & 43.18\\
\textbf{Kn} & 15.63 & \textbf{17.86} & 24.47 & 16.31 & 20.39 & \textbf{25.84} & 0.31 & \textbf{6.24} & 35.31 & \textbf{37.98}\\
\textbf{Ml} & 19.22 & 17.96 & 25.20 & 17.31 & 22.80 & \textbf{28.07} & 0.68 &\textbf{6.74} & 38.35 & \textbf{41.54}\\
\textbf{Mr} & 15.12 & \textbf{18.89} & 23.96 & 17.46 & 36.12 & 27.54 & 2.07 & \textbf{6.74} & 41.89 & 41.10\\
\textbf{Od} & 11.11 & \textbf{15.09} & 9.06 & \textbf{15.52} & 11.23 & \textbf{23.76} & 0.26 & \textbf{5.61} & 28.81 & \textbf{36.39} \\
\textbf{Pa} & 15.60 & \textbf{20.54} & 28.43 & 19.88 & 21.96 & \textbf{30.54} & 1.31 & \textbf{8.23} & 43.37 & \textbf{45.10}\\
\textbf{Ta} & 19.96 & 18.22 & 22.45 & 17.42 & 19.74 & \textbf{27.54} & 0.70 & \textbf{7.03} & 39.64 & \textbf{40.29}\\
\textbf{Te} & 16.07 & \textbf{17.79} & 23.93 & 17.72 & 13.59 & \textbf{25.37}& 0.53 & \textbf{6.38} & 34.97 & \textbf{39.24}\\
\textbf{En} & \multicolumn{2}{c}{27.41} & \multicolumn{2}{c}{27.13} & \multicolumn{2}{c}{34.69} & \multicolumn{2}{c}{8.55} & \multicolumn{2}{c}{52.17} \\
\bottomrule
\end{tabular}%
}
\caption{Performance of both Direct Inference and translate test inference of various Large Language Models on Zero-Shot Extractive INDIC QA BENCHMARK. We report the average numbers (F1) across span-extraction datasets.}
\label{ref:ext_temporary}
\end{table*}

\begin{figure*}[!h]
\centering
\includegraphics[width=1.0\linewidth]{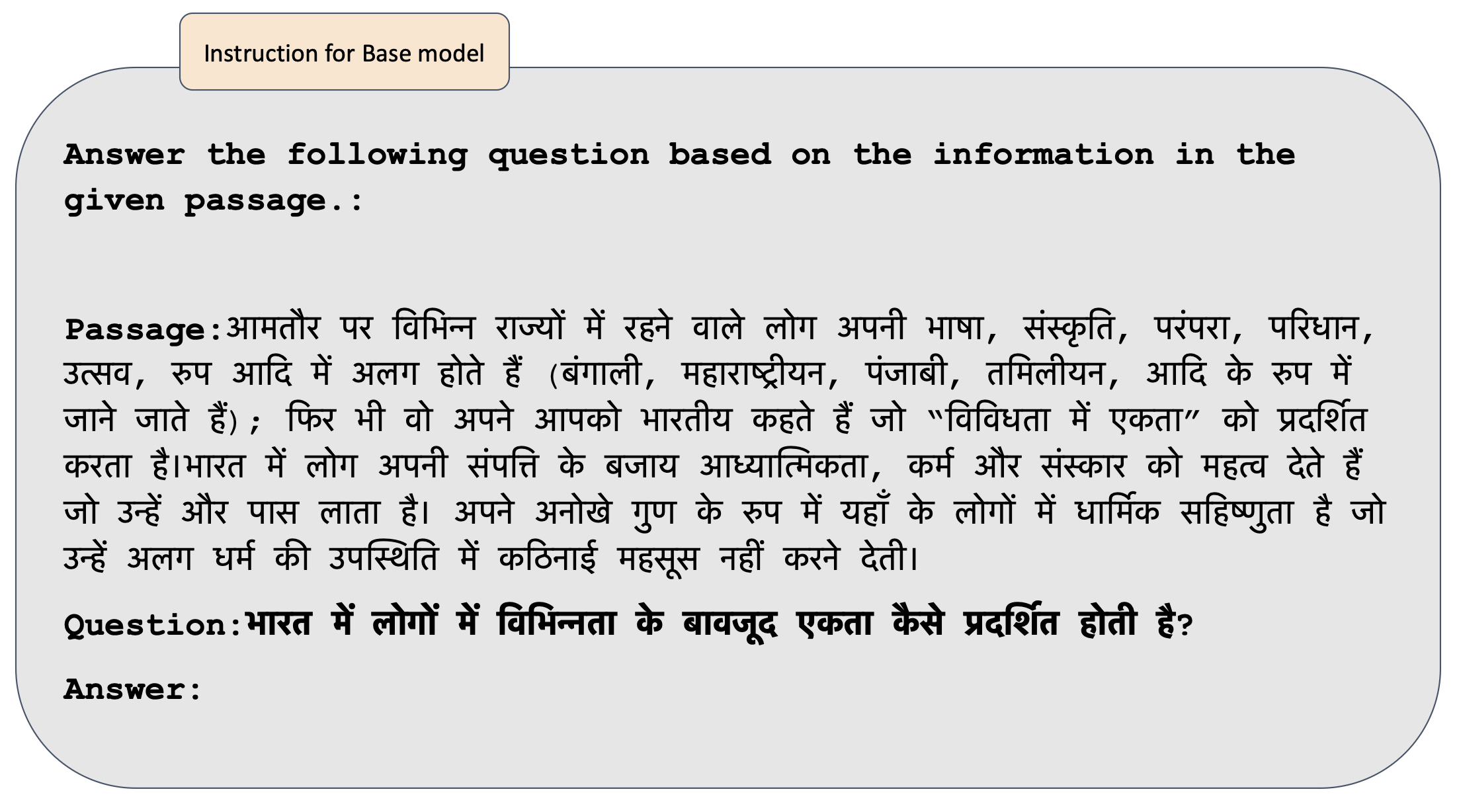}
\caption{Evaluation prompt used for the base model.}
\label{fig:prompt_to_base}
\end{figure*}

\begin{figure*}[!h]
\centering
\includegraphics[width=1.0\linewidth]{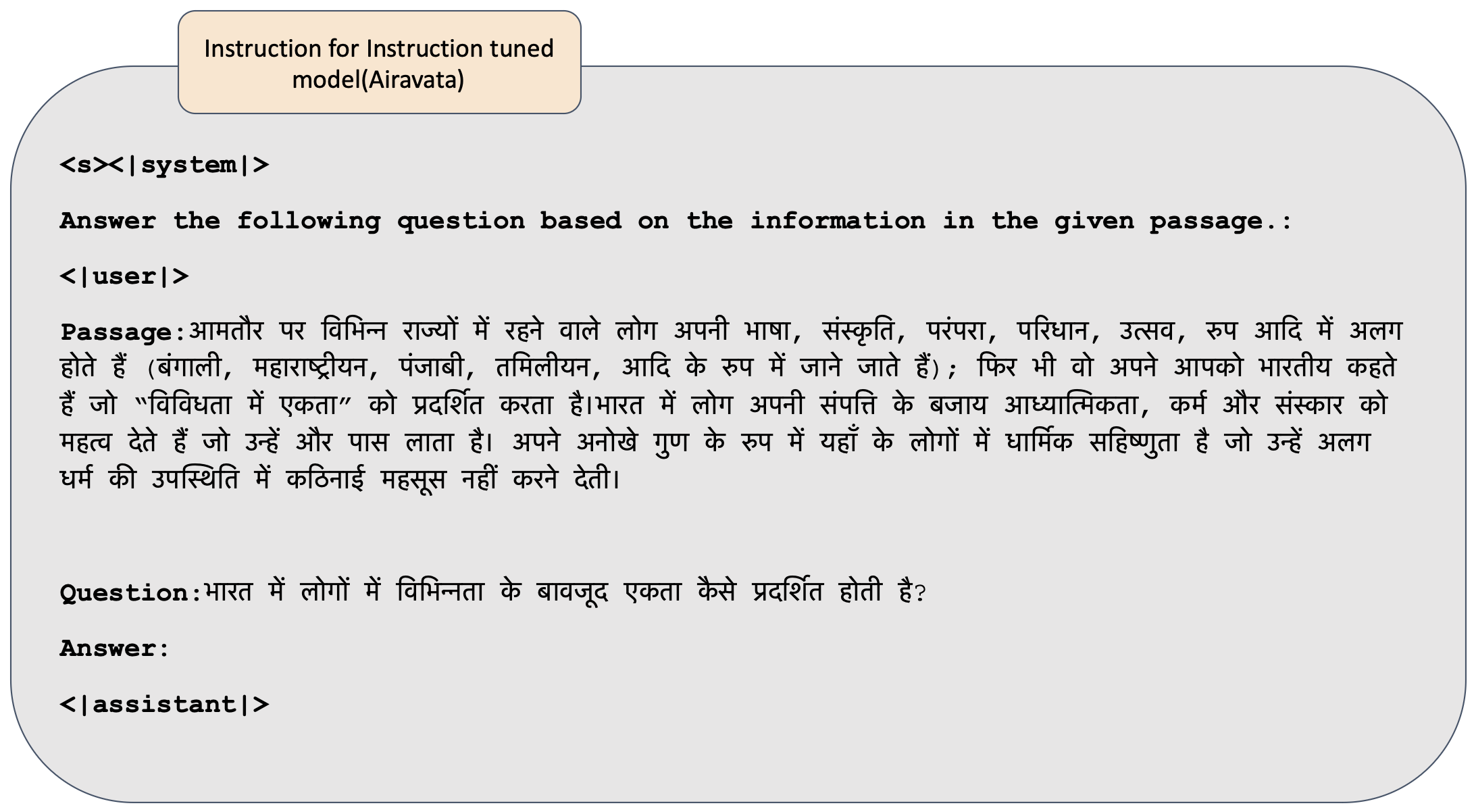}
\caption{Evaluation prompt used for the instruction-tuned model. Most models utilize a chat format. This is an example of the prompt used for the Airavata model.}
\label{fig:prompt_to_it}
\end{figure*}

\begin{figure*}[!h]
\centering
\includegraphics[width=1.0\linewidth]{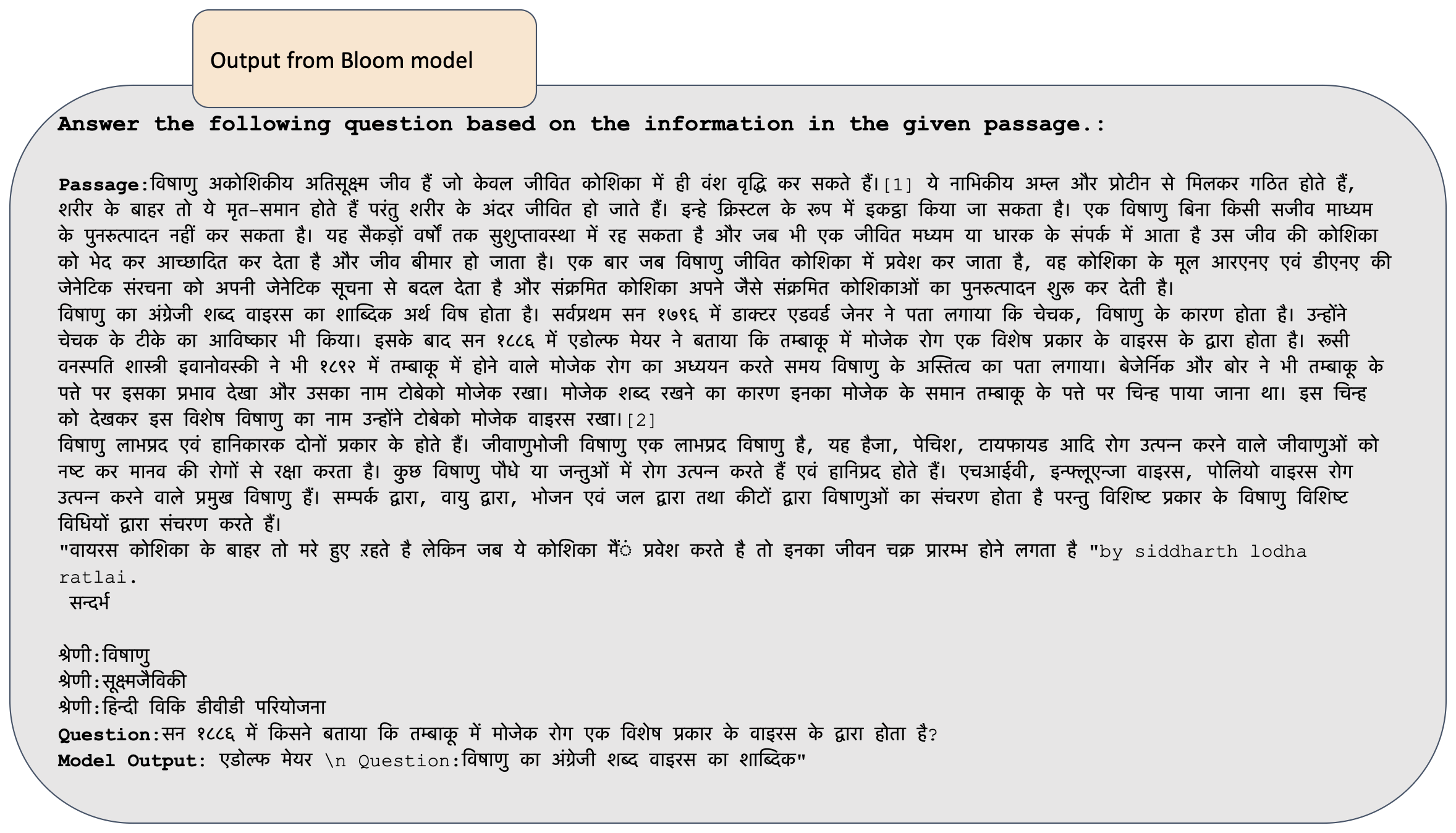}
\caption{Base Model(Bloom) prediction on one of the instance of extractive QA dataset, it is evident from the fig. the base model sometimes output garbage along with correct answer.}
\label{fig:Extractive_prediction_bloom}
\end{figure*}

\begin{figure*}[!h]
\centering
\includegraphics[width=1.0\linewidth]{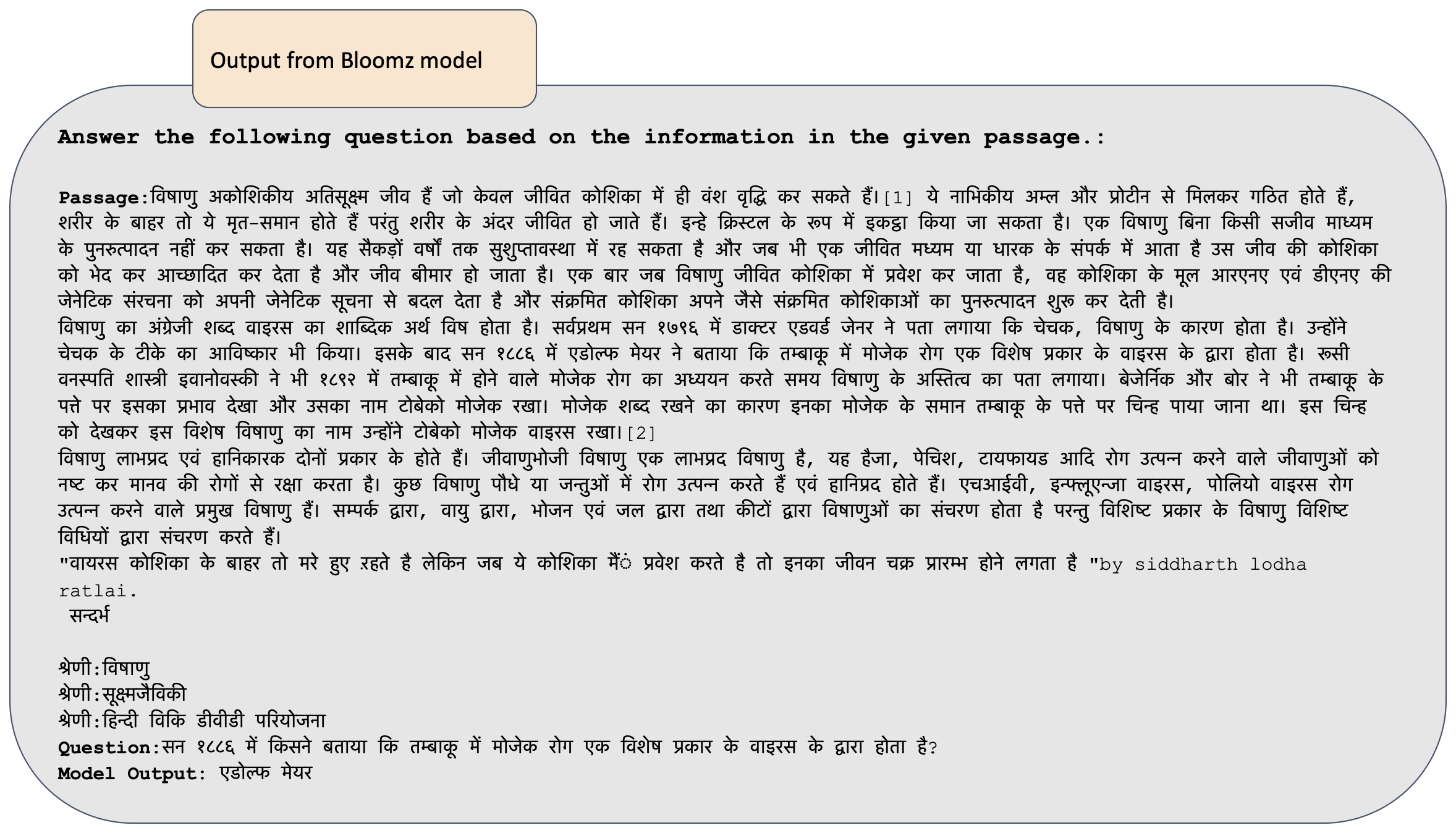}
\caption{Instruction tunned Model(Bloomz) prediction on one of the instance of extractive QA dataset.}
\label{fig:Extractive_prediction_bloomz}
\end{figure*}

\begin{figure*}[!h]
\centering
\includegraphics[width=1.0\linewidth]{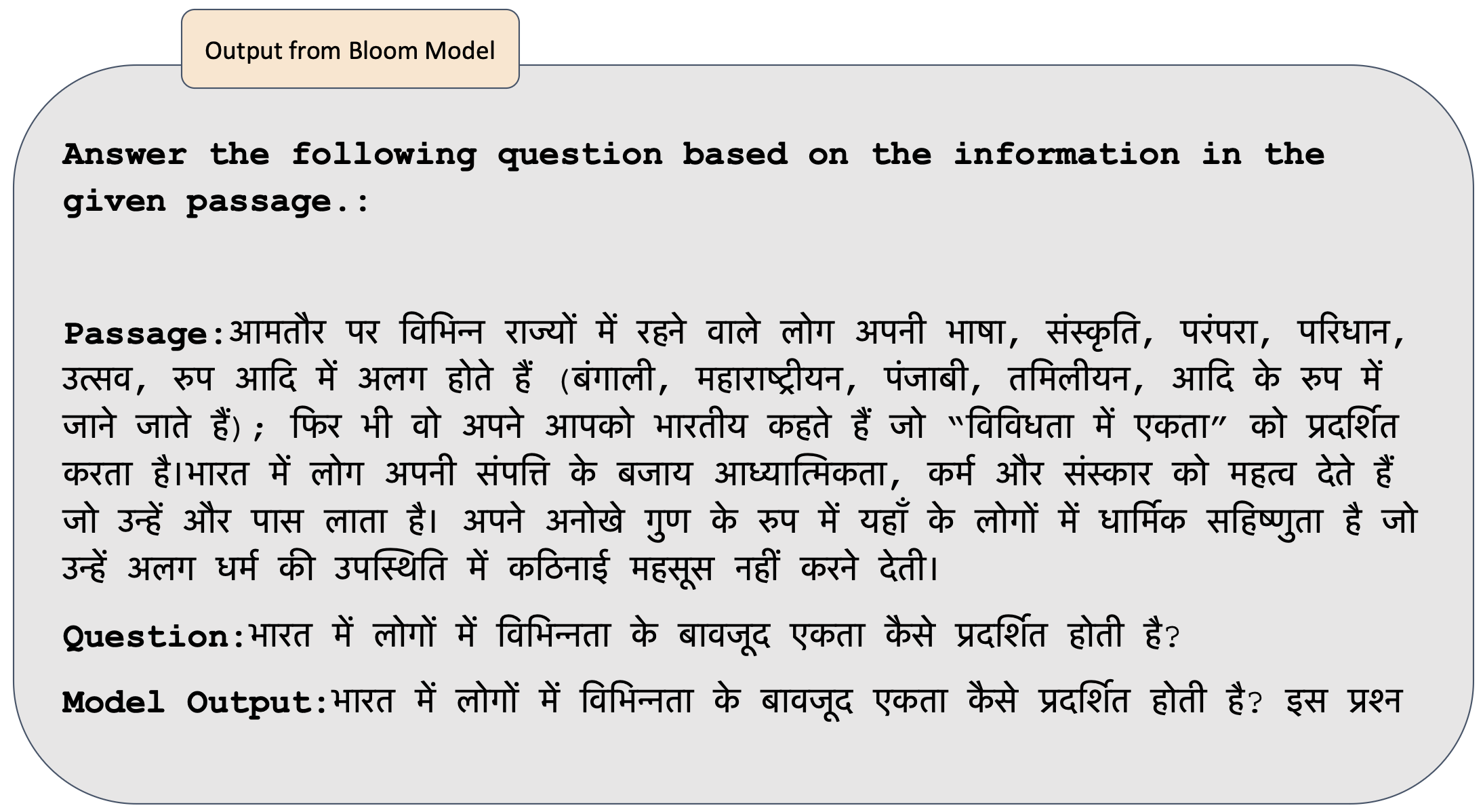}
\caption{Base Model(Bloom) prediction on one of the instance of Abstractive QA dataset}
\label{fig:Generative_prediction_bloom}
\end{figure*}

\begin{figure*}[!h]
\centering
\includegraphics[width=1.0\linewidth]{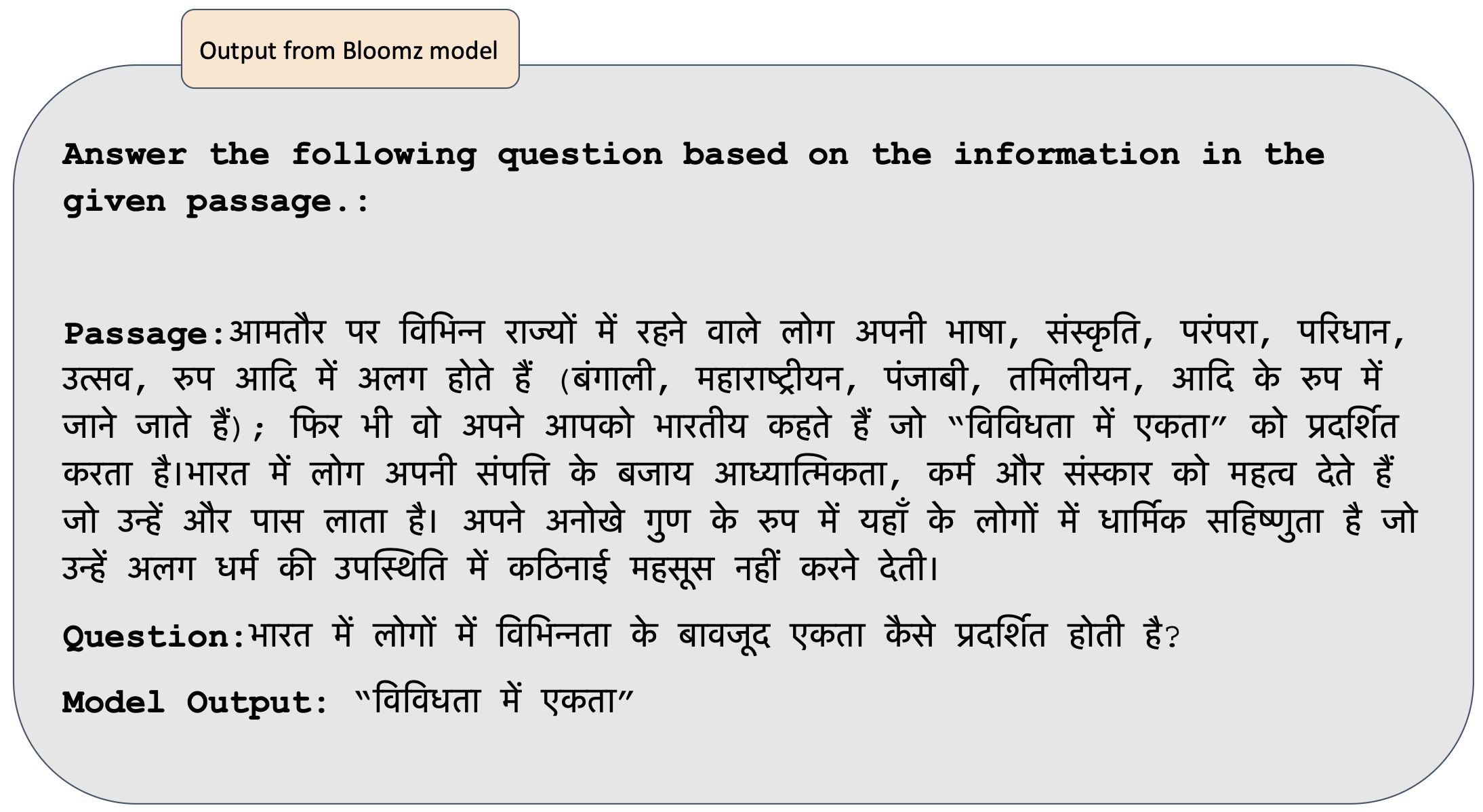}
\caption{Instruction tunned Model(Bloomz) prediction on one of the instance of Abstractive QA dataset.}
\label{fig:Generative_prediction_bloomz}
\end{figure*}

\begin{figure*}[h!]
\centering
\includegraphics[width=1.0\linewidth]{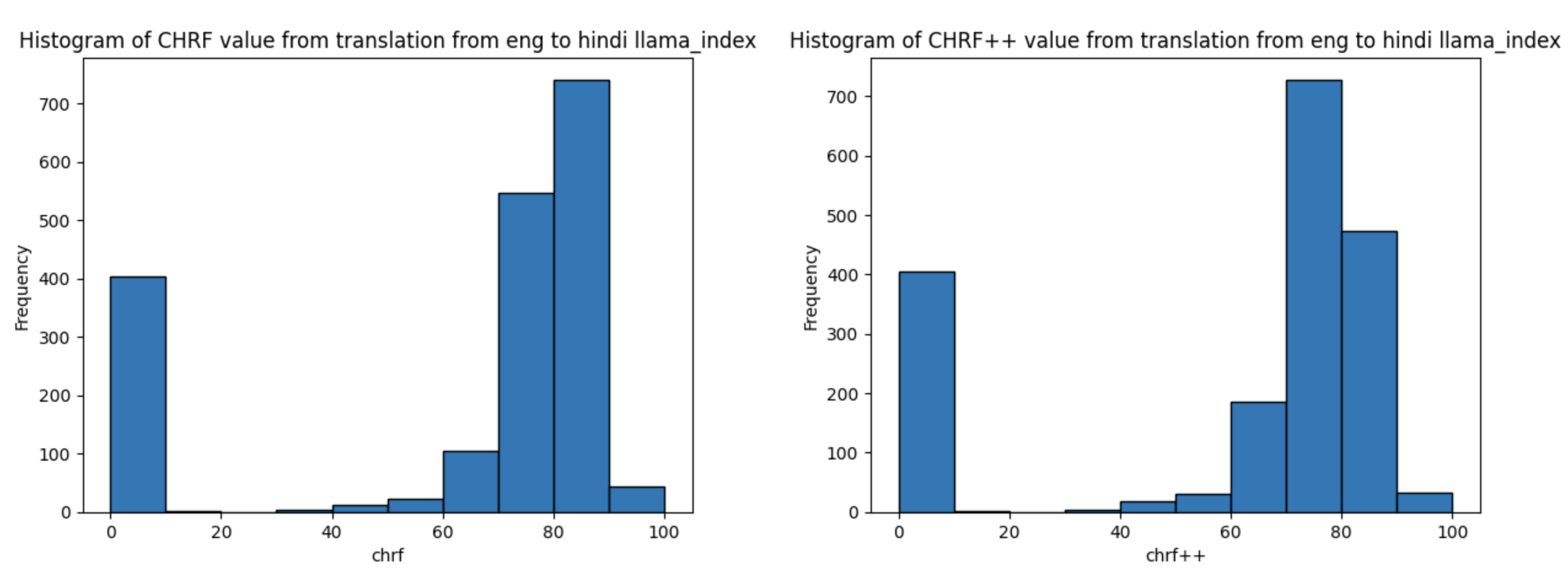}
\caption{}
\label{fig:llama_index_Trans}
\end{figure*}

\begin{figure*}[!h]
\centering
\includegraphics[width=1.0\linewidth]{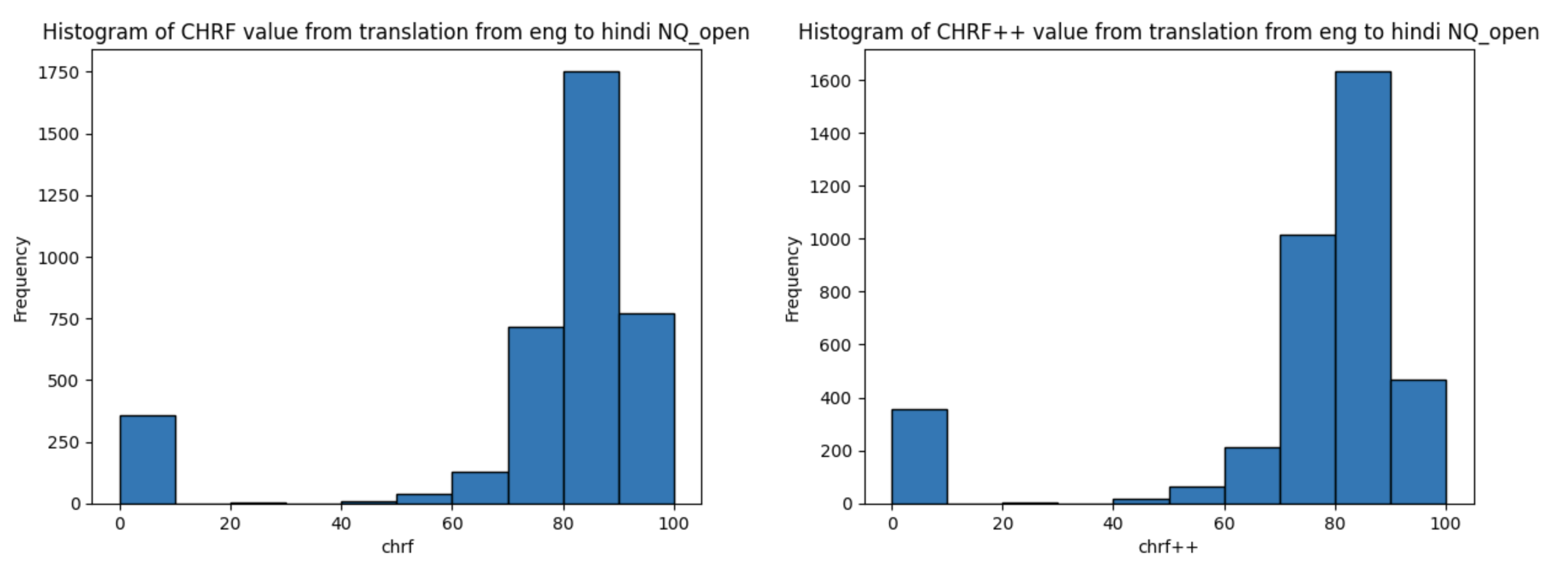}
\caption{CHRF and CHRF++ scores computed after translating the NQ open dataset from English to Hindi and then back-translating from Hindi to English. The scores are calculated between the original and back-translated parts, with a threshold of 50 applied to the CHRF scores to filter the data.}
\label{fig:NQ_open_Trans}
\end{figure*}

\begin{figure*}[h!]
\centering
\includegraphics[width=1.0\linewidth]{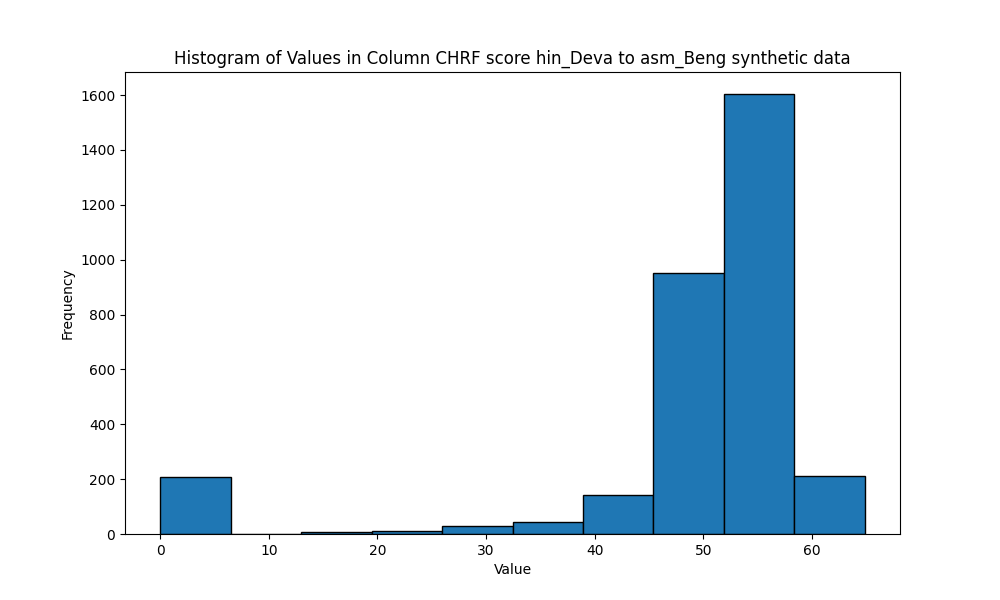}
\caption{CHRF scores computed after translating the synthetic dataset from  Hindi to asm and then back-translating from asm to Hindi.}
\label{fig:llama_index_Trans}
\end{figure*}

\begin{figure*}[h!]
\centering
\includegraphics[width=1.0\linewidth]{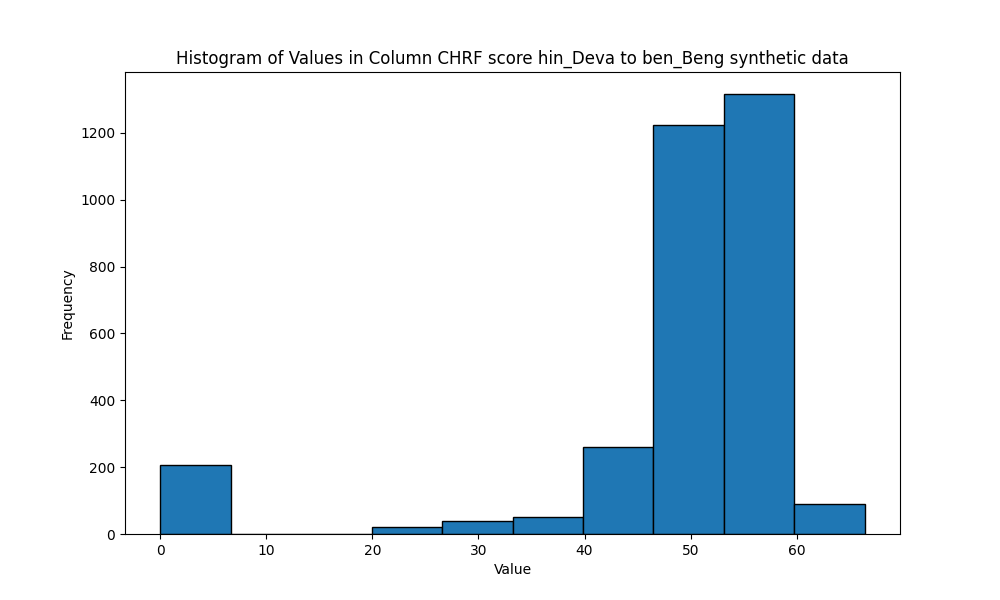}
\caption{CHRF and CHRF++ scores computed after translating the synthetic dataset from Hindi to ben and then back-translating from ben to Hindi.}
\label{fig:llama_index_Trans}
\end{figure*}

\begin{figure*}[h!]
\centering
\includegraphics[width=1.0\linewidth]{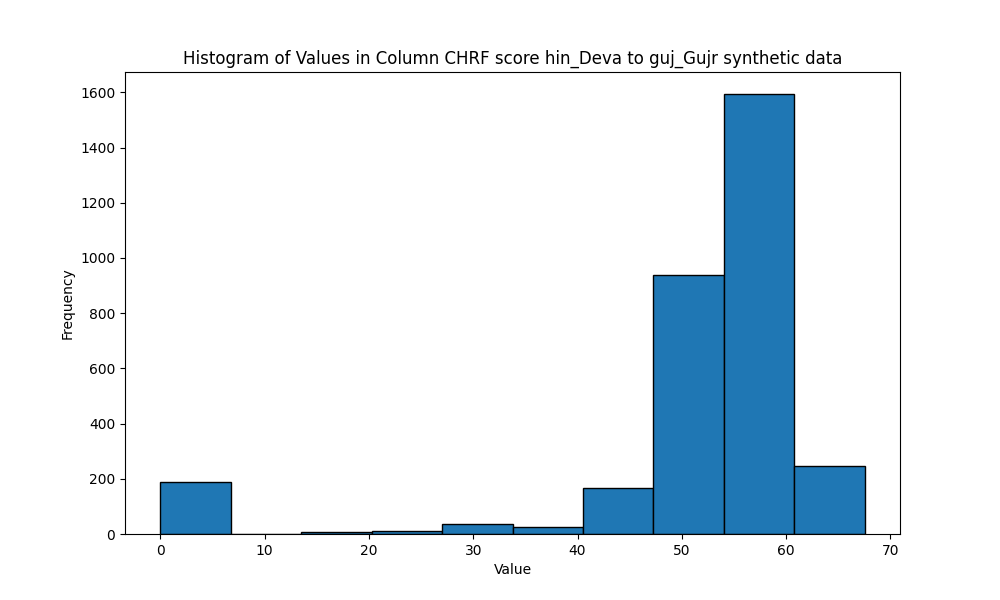}
\caption{CHRF and CHRF++ scores computed after translating the synthetic dataset from Hindi to guj and then back-translating from guj to Hindi.}
\label{fig:llama_index_Trans}
\end{figure*}

\begin{figure*}[h!]
\centering
\includegraphics[width=1.0\linewidth]{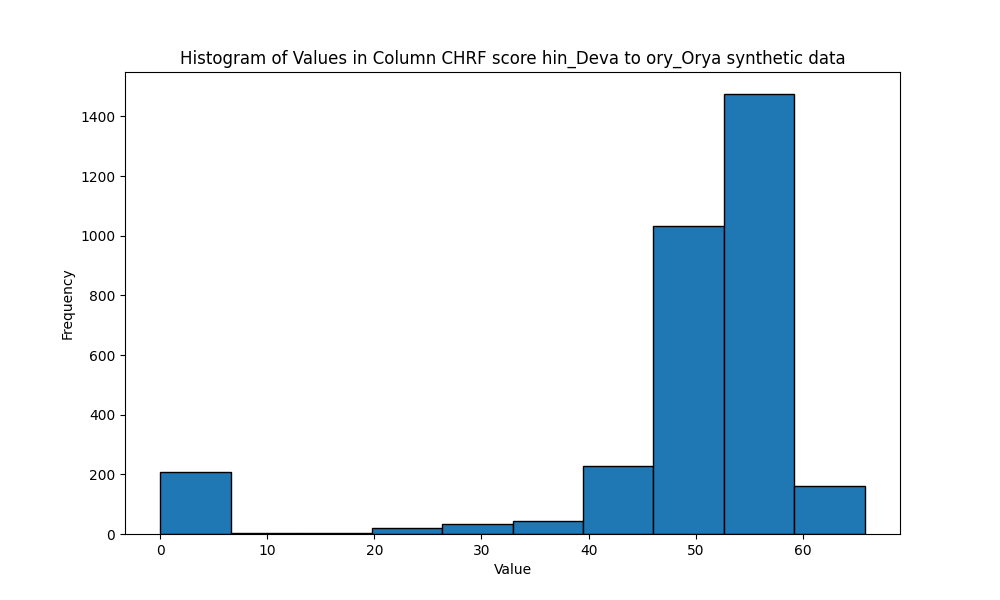}
\caption{CHRF and CHRF++ scores computed after translating the synthetic dataset from Hindi to ory and then back-translating from ory to Hindi.}
\label{fig:llama_index_Trans}
\end{figure*}

\begin{figure*}[h!]
\centering
\includegraphics[width=1.0\linewidth]{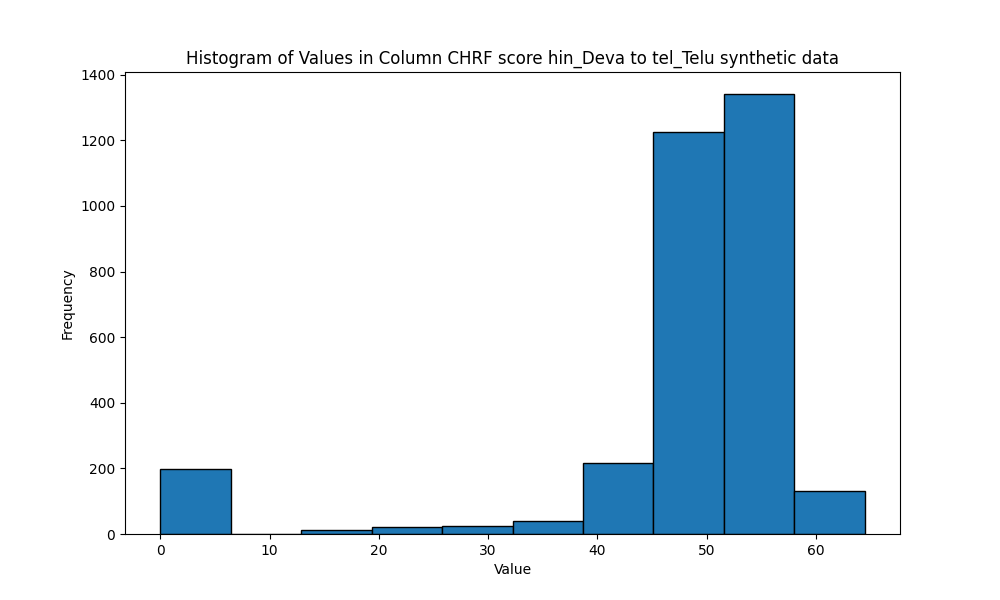}
\caption{CHRF and CHRF++ scores computed after translating the synthetic dataset from Hindi to kan and then back-translating from kan to Hindi.}
\label{fig:llama_index_Trans}
\end{figure*}

\section*{Frequently Asked Questions}

\subsection*{1) Why choose IndicTrans2 over other available translation models?}

Answer: \cite{gala2023indictrans2} show that IndicTrans2 surpasses other models, such as NLLB and Google Translate, particularly for English to Hindi translation tasks. In our analysis, we also tested NLLB, but the CHRF scores for back-translated and source texts were lower than those achieved with the IndicTrans2 model.


\end{document}